\documentclass[a4paper,oneside,reqno,10pt]{amsart}
\usepackage[latin1]{inputenc}
\usepackage{amsmath, amssymb,amsthm, mathtools, empheq}
\usepackage{paralist, mathtools}
\usepackage{caption}
\usepackage{subcaption}
\usepackage{graphicx}
\usepackage[colorlinks=true]{hyperref}
\hypersetup{linkcolor=blue, urlcolor=blue, citecolor=blue, filecolor=blue}
\usepackage{placeins}
\usepackage{enumerate}

\usepackage{xcolor}
\usepackage[normalem]{ulem}

\newtheorem{theorem}{Theorem}[section]

\newtheorem{lemma}[theorem]{Lemma}

\theoremstyle{definition}

\title[Feedforward Neural Network for Frequency Response]{A feedforward neural network for modeling of average pressure frequency response}


\makeatletter
\g@addto@macro{\endabstract}{\@setabstract}
\newcommand{\authorfootnotes}{\renewcommand\thefootnote{\@fnsymbol\c@footnote}}%
\makeatother

\date{August 8, 2021.}

\begin{document}
\maketitle

\begin{center}
\normalsize
\authorfootnotes
Klas Pettersson\textsuperscript{1}\footnote{Corresponding author; klaspe@chalmers.se.}, Andrei Karzhou\textsuperscript{2}, and Irina Pettersson\textsuperscript{3}
\par \bigskip

\textsuperscript{1} Chalmers University of Technology, Gothenburg, Sweden \par
\textsuperscript{2} University of Troms{\o}, Norway \par
\textsuperscript{3} Chalmers University of Technology and Gothenburg University, Sweden \par
\bigskip

%\today
\end{center}

%%%%%%%%%%%%%%%%%%%%%%%%%%%%%%%%%%%%%%%%%%%%%%%%%%%%%%%%%%

\begin{abstract}
The Helmholtz equation has been used for modeling the sound pressure field under a harmonic load.
Computing harmonic sound pressure fields by means of solving Helmholtz equation
can quickly become unfeasible if one wants to
study many different geometries for ranges of frequencies.
We propose a machine learning approach, namely a feedforward dense neural network, for computing the average sound pressure over a frequency range.
The data is generated with finite elements, by numerically computing the response of the average sound pressure,
by an eigenmode decomposition of the pressure.
We analyze the accuracy of the approximation and determine how much training data is needed in order to reach a certain
accuracy in the predictions of the average pressure response.
\end{abstract}

%\tableofcontents

\section{Introduction}

Modeling such acoustics problems as building acoustics, vehicle interior noise problems, noise reduction, insertion and transmission loss, often requires computing average sound pressure, which in its turn is based on the computation of natural frequencies and the response to a dynamic excitation.
There are two main approaches to modeling of acoustic systems
under small frequency excitation: To model acoustics in the time domain or in the frequency domain. Sound waves, as vibrations, are described by a time dependent wave equation, which can be reduced to a time independent Helmholtz equation by assuming harmonic dependence on time~\cite{skudrzyk2012foundations}. If the geometry of the domain is complex, it is decomposed into subdomains, in each subdomain the analysis is performed. Given an acoustic system, one needs to solve the Helmholtz equation repeatedly for given frequency, which might be very costly. In addition, the most important and costly part in a FEM-analysis is the computation of eigenfrequencies and eigenfunctions in each subdomain. We propose a feedforward dense neural network (multi-layer perceptron) for computing  the average sound pressure in cylindrical cavities with polygonal boundary.

Some motivation for studying the average pressure response over a range of frequencies 
can be found in engineering applications.
For example, standardized frequency ranges can be found in the ISO standard~\cite{iso20037235}.

For an overview of deep learning in neural networks, we refer to \cite{schmidhuber2015deep} and for overview of basic mathematical principles to \cite{goodfelow2016deep,strang2019linear} and literature therein. Application of machine learning methods in acoustics has made significant progress in recent years.
A comprehensive overview of the recent advances is given in \cite{bianco2019machine}. The frequency response problem, being the basis in modeling of acoustic problems, is not specifically addressed in \cite{bianco2019machine}, as any other combination of machine learning techniques and modeling with partial differential equations (PDEs).

There are many work devoted to solving PDEs, forward as well as inverse problems, by means of machine learning techniques. We mention just some of them.
The work \cite{lagaris1998artificial} propose an algorithm to solve initial and boundary value problems using artificial neural networks. The gradient descent is used for optimization. 
In \cite{ossandon2016neural} the authors propose an algorithm to solve an inverse problem associated with the calculation of the Dirichlet eigenvalues of the anisotropic Laplace operator. The finite elements are used to generate the training data. The main goal is to characterize the material properties (coefficient matrix) through the eigenvalues.

In \cite{sirignano2018dgm} the authors approximate solutions to high-dimensional PDEs with a deep neural network which is trained to satisfy the differential operator, initial condition, and boundary conditions. The convergence of the neural network to the solution of a PDE is proved. In contrast to \cite{lagaris1998artificial}, the algorithm in \cite{sirignano2018dgm} is mesh-free.

In \cite{adler2017solving} a partially learned approach is employed for the solution of ill-posed inverse problems. The paper contains also a good literature overview for inverse problems. In \cite{berg2018unified} the authors propose deep feedforward artificial neural networks (mesh-free) to approximate solutions to partial differential equations in complex geometries. 
The paper \cite{raissi2019physics} focuses on the nonlinear partial differential equations. 
Some other network inspired approaches in the study of PDEs are \cite{baymani2011feed, baymani2015artificial, ossandon2016neural-Lame, ossandon2017neural}.

In \cite{rizzuti2019learned} the authors propose an iterative solver for the Helmholtz equation which combines traditional Krylov-based solvers with machine learning. The result is a reduced computational complexity.

In the present paper we use feedforward fully connected neural networks with the ReLU activation function in hidden layers in order to approximate the average pressure function originated in frequency response problems. We choose to use three hidden layers, 128 nodes in each, and ADAM optimizer. The step size in the gradient descent is scheduled to have polynomial decay. A more detail description of the neural network is provided in Section \ref{sec:dnn}.

The feedforward neural network is designed to directly learn the average pressure, in contrast to the works cited above, where the neural networks are tailored to predict the coefficients of the inverse problem or solutions to PDEs.

It is known that a neural network can approximate any continuous function to an arbitrary accuracy \cite{hornik1990universal}. We focus on the frequency response problem (low frequencies) in two-dimensional polygonal cylinders. Assuming harmonic load on a part on the boundary, we arrive at a time independent Helmholtz equation for the sound pressure. The mean-value of the average pressure over a given frequency range is an important quantity for characterizing the sound attenuation, insertion and transmission losses. The numerical solution of this problems implies solving the Helmholtz equation for many different values of the spectral parameter, which is a costly problem. Besides, the pressure function is singular near the eigenvalues of the Laplace equation, and the standard quadrature schemes cannot be applied in order to compute the average $\Psi$ over a frequency range (see Section~\ref{sec:num-psi} and Figure~\ref{fig:uninum}(b) for the explanation). Instead, we represent the average pressure $\Psi$ (objective function) in terms of a Hilbert basis, the eigenmodes of the Laplace operator. We generate data sets containing around $700\,000$ randomly generated points which define polygonal cylinders and the corresponding objective functions $\Psi$ computed using finite elements.
A feedforward neural network with five input nodes (coordinates defining cylinders), three hidden layers and one output node (scalar objective function $\Psi_{\mathrm{ml}}$) is then constructed to approximate the objective function.

We analyze the performance of the model, and show the dependency of the mean squared error (MSE) on the training set size.
Moreover, we analyze how many samples is needed to reach a desired approximation accuracy.
For example, for polygonal cylinders defined by five randomly generated points,
on average over $95\%$ are predicted with mean absolute error less than $0.01$
when the training set contains $200\,000$ data points. 
The data used for machine learning in this paper is available at~\cite{Pettersson2020}.

The sound pressure as a function of frequency is nonlinear, and thus the linear regression methods perform poorly.
In Section~\ref{sec:lin-vs-nonlin}, we show the results of the approximation of the objective function by means of linear regression vs. feedforward fully connected neural network with ReLU nonlinearity. The proposed method performs much better, as expected.

For machine learning we have used  Tensorflow~\cite{tensorflow2015-whitepaper}, and
the stochastic gradient descent optimizer %JOGI~\cite{yogi} and 
ADAM~\cite{kingma2014adam}.
For the numerical computation of the average pressure we used primarily the SLEPc~\cite{hernandez2005slepc},
with user interfaces and numerical PDE tools FreeFem~\cite{MR3043640} and FEniCS~\cite{AlnaesBlechta2015a} to the standard 
numerical packages.

Our method can be applied for analyzing frequency response in elastic bodies and fluid-structure interaction problems.
In three-dimensional case the frequency response problems become computationally heavy, and the effectiveness of the stochastic gradient descent gives some hope for significant reduction of data needed for training.

The rest of this paper is organized as follows.
In Section~\ref{sec:frequencyresponse}, the numerical method for computing the average sound pressure
response is described.
Using the numerical method, the data sets for polygonal cylinders are generated 
and the data sets are described in Section~\ref{sec:datasets}.
In Section~\ref{sec:dnn}, we specify the feedforward dense neural network
and the choices of for the training procedure.
The model obtained after training is evaluated in Section~\ref{sec:evaluation},
and compared to a linear model in Section~\ref{sec:lin-vs-nonlin}.

\section{Frequency response problem and average pressure}
\label{sec:frequencyresponse}

Assume that a domain $\Omega$ is occupied by a inviscid, homogeneous, compressible fluid (liquid or gas). There are several options for choosing a primary variable
for small amplitude vibrations: fluid displacement, acoustic pressure or fluid velocity potential.
We are going to use a description in terms of the scalar pressure function $P$.
Let $c$ be the speed of sound in the fluid,
and $\rho$ be the mass density of the fluid, both assumed not to depend on the pressure $P$.
The equation of motion without taking in account the damping is the wave equation for the acoustic pressure \cite{skudrzyk2012foundations}:
\begin{align}
    \label{eq:wave}
    \rho \frac{\partial^2 P(t,x)}{\partial t^2} - c^2 \Delta P(t,x) = F(t,x).
\end{align}
Here $F$ is the applied load.
The solution of the last equation is by linearity a sum of a particular solution to a non-homogeneous equation (forced motion) and the general solution of the homogeneous equation (natural motion). If the excitation is harmonic $F(t,x)=f(x)\cos(\omega t) = f(x) \Re e^{i\omega t}$, the forced motion is called the steady-state response.
The real-valued pressure together with the phase angle is then called the dynamic frequency response of the system \cite{choi2006structural}.
To eliminate the time dependency in the wave equation, we substitute $P(t,x)=\Re (p(x)e^{i\omega t})$ into it and obtain a time independent Helmholtz equation for the amplitude $p$:
\begin{align*}
    -\Delta p(x) - \frac{\omega^2 \rho}{c^2} p(x) & = f(x).
\end{align*}
If the acoustic medium is contained in a bounded domain $\Omega$, we will need to impose boundary conditions on the boundary $\partial \Omega$. We will impose a harmonic load $\cos(\omega t)$ on the part of the boundary $\Gamma_D$, which results in the non-homogeneous Dirichlet boundary condition $p=1$.
On the rest of the boundary $\Gamma_N=\partial \Omega \setminus \Gamma_D$ is assumed to be sound hard (zero-flux condition). The problem in the frequency domain takes the form
\begin{align}\label{eq:FR-orig}
-\Delta p - \frac{\omega^2 \rho}{c^2} p & = 0 \quad \, \text{ in } \Omega, \notag\\
p & = 1 \quad\, \text{ on } \Gamma_D, \\
\nabla p \cdot \nu & = 0 \quad\, \text{ on } \Gamma_N,\notag
\end{align}
where $\nu$ is the exterior unit normal.

We are going to solve \eqref{eq:FR-orig} analytically for cylinders $\Omega$ with constant and non-constant cross-section.
We will obtain expressions for the mean-value of the solution to \eqref{eq:FR-orig} with respect to the spatial variable $\langle p \rangle=|\Omega|^{-1}\int_\Omega p \,dx$,  and its average with respect to the spectral parameter 
\begin{align*}
\lambda &= \frac{\omega^2\rho}{c^2}.\\
\end{align*}
The domain may be an open set in Euclidean space $\mathbf{R}^n$.
In what follows, we will work with $\Omega$ a bounded Lipschitz domain in $\mathbf{R}^2$,
that is a bounded open connected subset of $\mathbf{R}^2$ with Lipschitz continuous boundary.
Specifically, $\Omega$ will be a finite cylinder with polygonal boundary.

We will in the sequel assume the quantities and variables to be scaled
in such a way they are nondimensionalized, and thereby also suppress units from both manipulations and figures.

%%%%%%%%%%%%%%%%%%%%%%%%%%%%%%%%%%%%%%%%%%%%%%%%%%%%%%%%%%%
\subsection{Uniform cylinders}

We start with cylinders with constant cross-section, where one can find explicit formulas for eigenfunctions and eigenvalues for the Laplace operator and therefore solve the frequency response problem analytically.

Let us denote $\Omega=(0,1) \times (-a,a)$ for $r_{\min} \le a \le r_{\max}$ a uniform cylinder.
The boundary of $\Omega$ consists of two parts, and we denote $\Gamma_D=\{(x_1,x_2):\,\, x_1=0\}$ (the part where a Dirichlet boundary condition will be imposed) and $\Gamma_N=\partial \Omega \setminus \Gamma_D$ (with a Neumann boundary condition). Consider the frequency response problem~\eqref{eq:FR-orig} in $\Omega$.
By the Fredholm alternative, there exists a unique solution $p_\lambda \in H^1( \Omega )$ to~\eqref{eq:FR-orig} if and only if $\lambda$ is not an eigenvalue of the Laplace operator in the cylinder:
\begin{align}\label{eq:evpuni}
-\Delta \psi & = \lambda \psi \quad\, \text{ in } \Omega, \notag\\
\psi & = 0 \quad\,\,\,\, \text{ on } \Gamma_D, \\
\nabla \psi \cdot \nu & = 0 \quad\,\,\,\, \text{ on } \Gamma_N.\notag
\end{align}
By the Hilbert-Schmidt and the Riesz-Schauder theorems,
the spectrum of~\eqref{eq:evpuni} is positive, discrete, countably infinite,
and each eigenvalue of finite multiplicity,
\begin{align*}
0 < \lambda_1 < \lambda_2 \le \lambda_3 \le \cdots \le \lambda_n \to \infty, \quad n \to \infty.
\end{align*}
Moreover, the eigenfunctions $\psi_i$ form an orthonormal basis under a proper normalization.
By separation of variables, choosing a convenient enumeration,
the eigenvalues $\lambda_{i,k,l}$ to~\eqref{eq:evpuni} are given by
\begin{align}
\label{eq:lambda-uni}
\lambda_{i,k,l} = \mu_k + \eta_{i,l},\quad i = 1,2, \quad k,l = 0,1,\ldots,
\end{align}
where
\begin{align}
\label{eq:mu-eta-uni}
\mu_k & = \frac{(2k+1)^2 \pi^2}{4}, \quad k = 0,1, \ldots,\\
\eta_{1,l} & = \frac{l^2 \pi^2}{a^2}, \quad
\eta_{2,l} = \frac{(2l+1)^2\pi^2}{4a^2}, \quad l = 0,1, \ldots.
\end{align}
The sequences $\mu_k > 0$ and $\eta_{i,l} \ge 0$
are the Dirichlet-Neumann eigenvalues of the Laplace operator on $(0,1)$,
and the Neumann eigenvalues of the Laplace operator on $(-a,a)$, respectively.
The eigenfunctions $\psi_{i,k,l}$ to \eqref{eq:evpuni} corresponding to the 
eigenvalues $\lambda_{i,k,l}$ are
\begin{align}
\label{eq:psi-uni}
\psi_{1,k,l} & = a_{1,k,l}\sin(\sqrt{\mu_k}x_1)\cos(\sqrt{\eta_{1,l}}x_2), \\ 
\psi_{2,k,l}&  = a_{2,k,l}\sin(\sqrt{\mu_k}x_1)\sin(\sqrt{\eta_{2,l}}x_2), \nonumber
\end{align}
where $a_{i,k,l}$ are $L^2(\Omega)$ normalization factors defined by
\begin{align*}
\int_{\Omega} \psi_{i,k,l} \psi_{j,p,q} \,dx & =
\begin{cases}
1 & \text{ if } (i,k,l) = (j,p,q),\\
0 & \text{ otherwise.}
\end{cases}
\end{align*}
Explicitly, $a_{1,k,0} = \sqrt{1/a}$, and otherwise $a_{i,k,l} = \sqrt{2/a}$.

Let $\lambda \in \mathbf{R}$ not be an eigenvalue to~\eqref{eq:evpuni}.
Then the method of separation of variables gives a solution $p_\lambda$ to~\eqref{eq:FR-orig} in the case of uniform cylinder:
\begin{align}\label{eq:plambdauni}
p_\lambda 
& = 1 + \sum_{k = 0}^\infty \frac{\lambda}{\mu_k - \lambda}
\frac{2}{\sqrt{\mu_k}} \sin( \sqrt{\mu_k} x_1 ) \notag\\
& = \cos(\sqrt{\lambda}x_1) + \tan(\sqrt{\lambda})\sin(\sqrt{\lambda}x_1).
\end{align}
Remark that $p_\lambda$ is constant in the $x_2$-direction for this particular choice of harmonic load $p(0,x_2)=1$.

\begin{figure}[hb!]
    \centering
    \begin{minipage}{.49\textwidth}
    \centering
    \includegraphics[width=.99\linewidth]{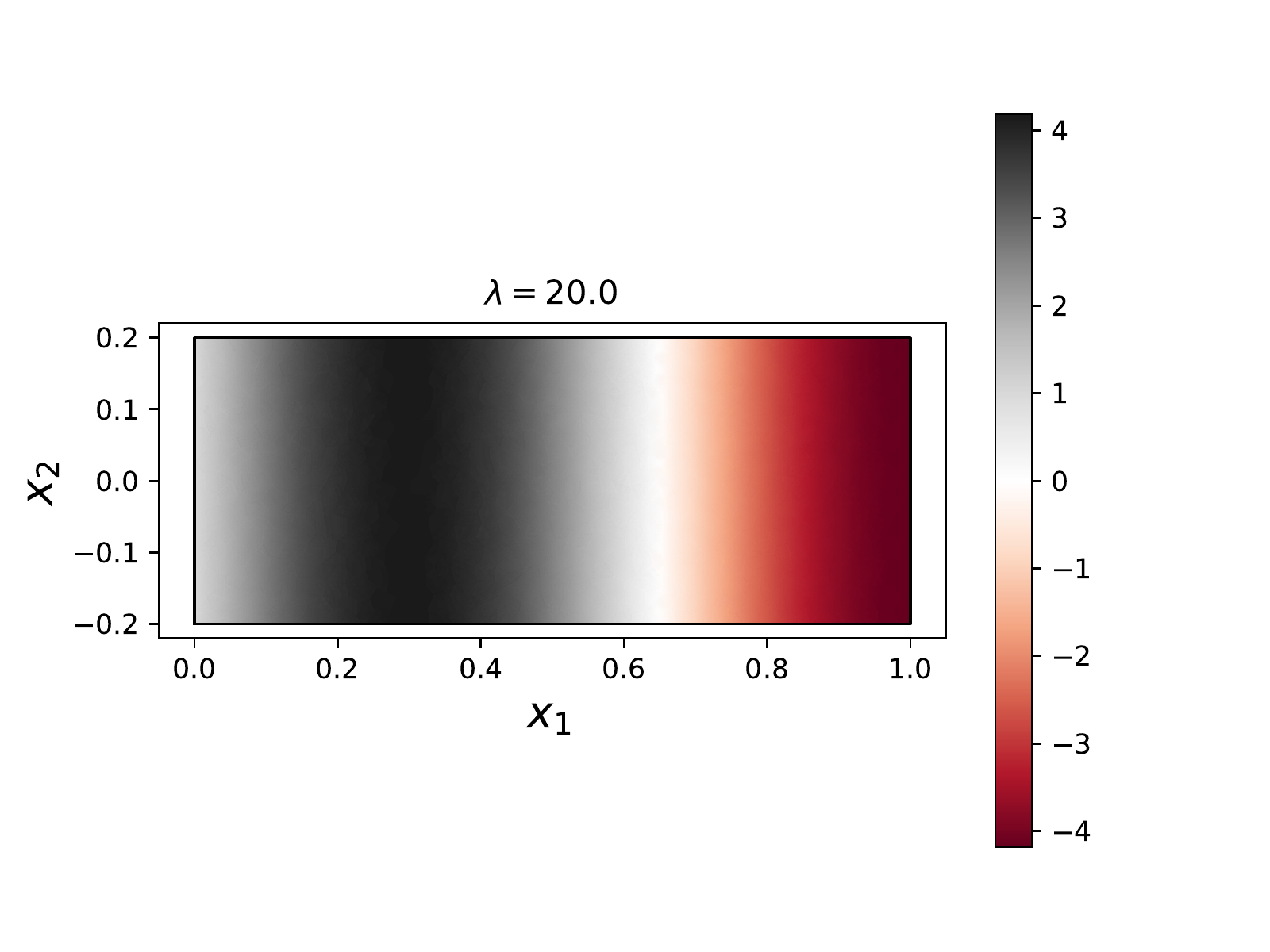}\\
    (a)
    \end{minipage}
    \begin{minipage}{.49\textwidth}
    \centering
    \includegraphics[width=.99\linewidth]{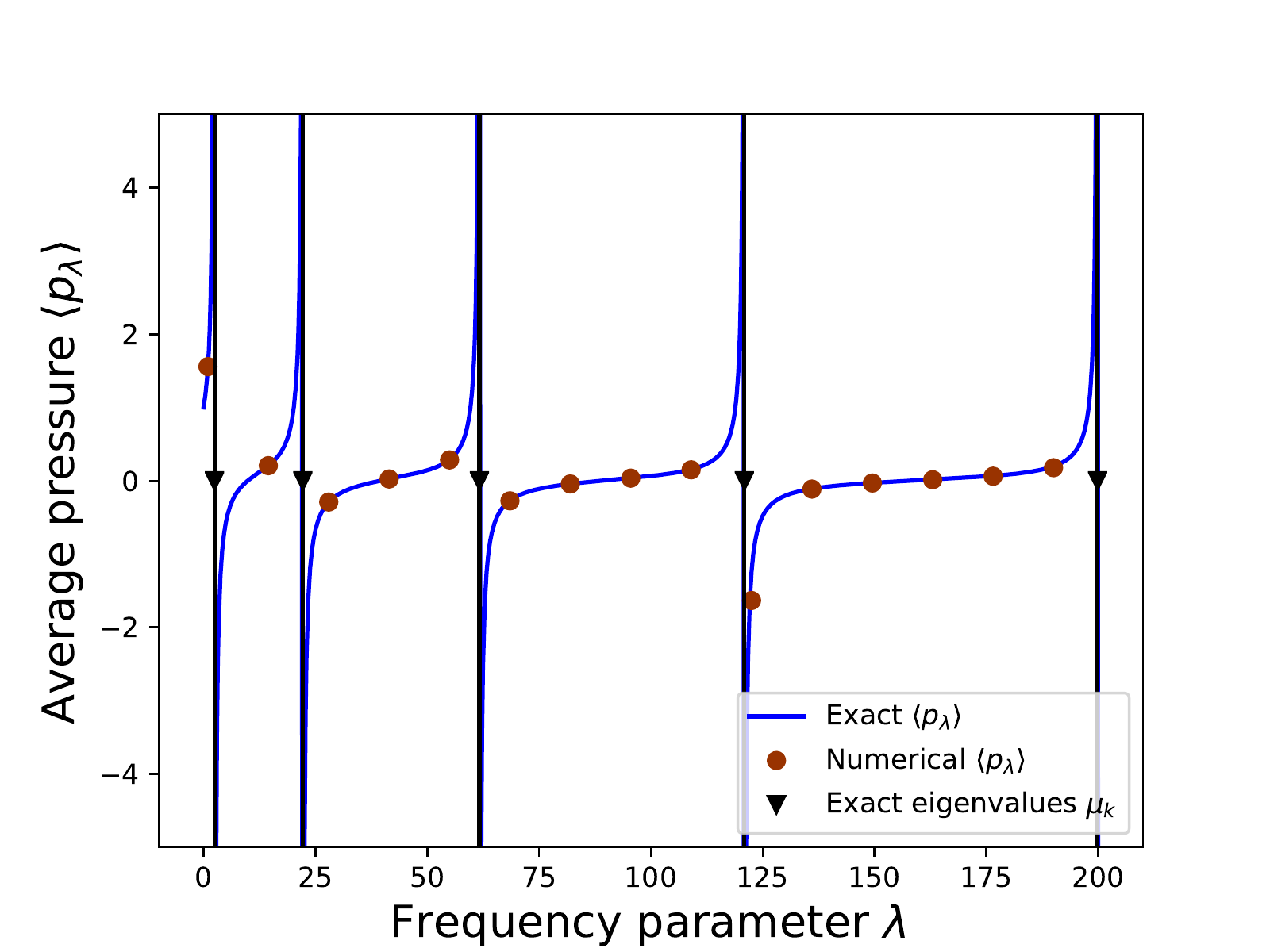}\\
    (b)
    \end{minipage}
    \caption{The response $p_\lambda$ for $\lambda = 20$ (a), and the average pressure
             $\langle p_\lambda \rangle$ (b)
             for a uniform cylinder.}
    \label{fig:uniresponse}
\end{figure}

When analyzing the acoustic response, one could be interested in the average pressure defined for a frequency sweep, namely the average of $p_\lambda$ with respect to $x$ and $\lambda$. 
Let us first compute the average pressure $\langle p_\lambda \rangle$ with respect to $x$:
\begin{align}\label{eq:avgp}
\langle p_\lambda \rangle & = \frac{1}{|\Omega|}\int_\Omega p_\lambda \,dx 
=
\begin{cases}
\displaystyle \frac{\tan (\sqrt{\lambda})}{\sqrt{\lambda}} & \text{ if } \lambda > 0,\\
1 & \text{ if } \lambda = 0.
\end{cases}
\end{align}
The pressure $p_\lambda$ and its mean-value with respect to the space variable as a function of $\lambda$ is shown in Figure~\ref{fig:uniresponse}.

The form of the response $\langle p_\lambda\rangle $ in \eqref{eq:avgp}
 indicates that it could be challenging to numerically evaluate an integral of $\langle p_\lambda\rangle $ in $\lambda$
over an interval $(\lambda_{\mathrm{min}}, \lambda_{\mathrm{max}})$ that contains a pole, because the computation of the Cauchy principal value of the integral requires both the location of the poles and their orders.
For the uniform cylinder we obtain the following explicit formula for the objective function:
\begin{align}
\Psi & = \frac{1}{\lambda_{\mathrm{max}} - \lambda_{\mathrm{min}}} \, \mathrm{p.v.}  \int_{\lambda_{\mathrm{min}}}^{\lambda_{\mathrm{max}}}
 \langle p_\lambda \rangle \,d\lambda \nonumber \\
 & = \frac{1}{\lambda_{\mathrm{max}} - \lambda_{\mathrm{min}}} \, \mathrm{p.v.}  \int_{\lambda_{\mathrm{min}}}^{\lambda_{\mathrm{max}}}
 \frac{\tan(\sqrt{\lambda})}{\sqrt{\lambda}} \, d\lambda \notag\\
 %& = \frac{1}{\lambda_{\mathrm{max}} - \lambda_{\mathrm{min}}}
 %( -2\log |\cos(\sqrt{\lambda_{\max}})| + 2\log |\cos(\sqrt{\lambda_{\min}})|  ) \\
 & = \frac{2}{\lambda_{\mathrm{max}} - \lambda_{\mathrm{min}}}
 \log \left| \frac{ \cos(\sqrt{\lambda_{\min}}) }{ \cos(\sqrt{\lambda_{\max}}) } \right|,\label{eq:psi-exact}
\end{align}
which is defined as long as both $\lambda_{\min}, \lambda_{\max}$ are not eigenvalues of~\eqref{eq:evpuni}.
More precisely, 
if $\lambda = \lambda_{i,k,l}$ is an eigenvalue to~\eqref{eq:evpuni},
the response
$p_\lambda$ exists if and only if $\int_\Omega \psi_{i,k,l} \,dx = 0$
for all eigenfunctions corresponding to $\lambda_{i,k,l}$,
by the Fredholm alternative.
One notes that $\int_\Omega \psi_{1,k,l} \,dx = 0$ for $l \ge 1$,
and $\int_\Omega \psi_{2,k,l} \,dx = 0$ for $l \ge 0$.
Thus for $\lambda = \lambda_{1,k,l}$ with $l \ge 1$, and for 
$\lambda = \lambda_{2,k,l}$ with $l \ge 0$, the solution $p_\lambda$
is unique modulo a linear combination of the corresponding eigenfunctions.
Such eigenfunctions do not contribute to the mean-value $\langle p_\lambda \rangle$ and therefore also not to $\Psi$.
It follows that~\eqref{eq:avgp} holds for $\lambda \neq \mu_k$,
and \eqref{eq:psi-exact} holds for $\lambda_{\min}, \lambda_{\max} \neq \mu_k$.

%\clearpage 

%%%%%%%%%%%%%%%%%%%%%%%%%%%%%%%%%%%%%%%%%%%%%%%%%%%%%%%%%
\subsection{Cylinders with varying cross-section}
\label{sec:nonuniform}

In this section we will show how a Hilbert basis can be used  to compute the average in $\lambda$ of $\langle p_\lambda \rangle$ (the objective function) in the case when cylinders have varying cross-section. The explicit formulas for the eigenvalues and eigenfunctions like \eqref{eq:lambda-uni}--\eqref{eq:psi-uni} are not available any more, and we will use the finite elements to compute the eigenpairs of the Laplace operator.

Let $p$, as before, for a given $\lambda$ solve the frequency response problem
\begin{align}\label{eq:FR-nonuni}
-\Delta p - \lambda p & = 0 \quad \, \text{ in } \Omega, \notag\\
p & = 1 \quad\, \text{ on } \Gamma_D, \\
\nabla p \cdot \nu & = 0 \quad\, \text{ on } \Gamma_N.\notag
\end{align}
The cylinder $\Omega$ is not uniform any more, and can be described by $\Omega= \{x=(x_1, x_2): \,\, x_1\in (0,1), \,\, x_2 \in I(x_1)\}$, where $I(x_1)=(-a(x_1), a(x_1))$ is an interval such that $r_{\text{min}}\le a(x_1)\le r_{\text{max}}$.

\begin{figure}[hb!]
    \centering
    \begin{minipage}{.49\textwidth}
    \centering
    \includegraphics[width=.99\linewidth]{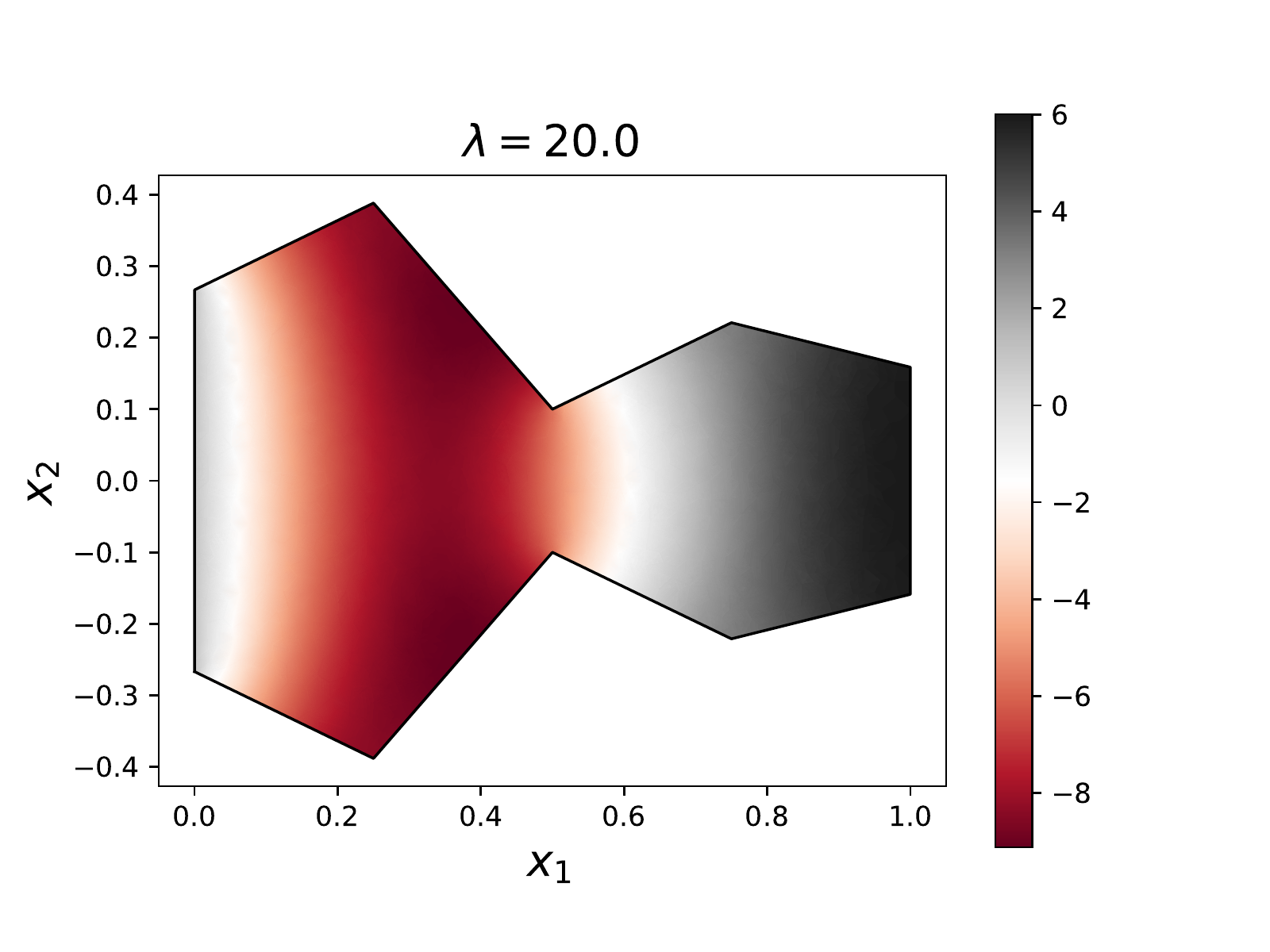}
    (a)
    \end{minipage}
    \begin{minipage}{.49\textwidth}
    \centering
    \includegraphics[width=.99\linewidth]{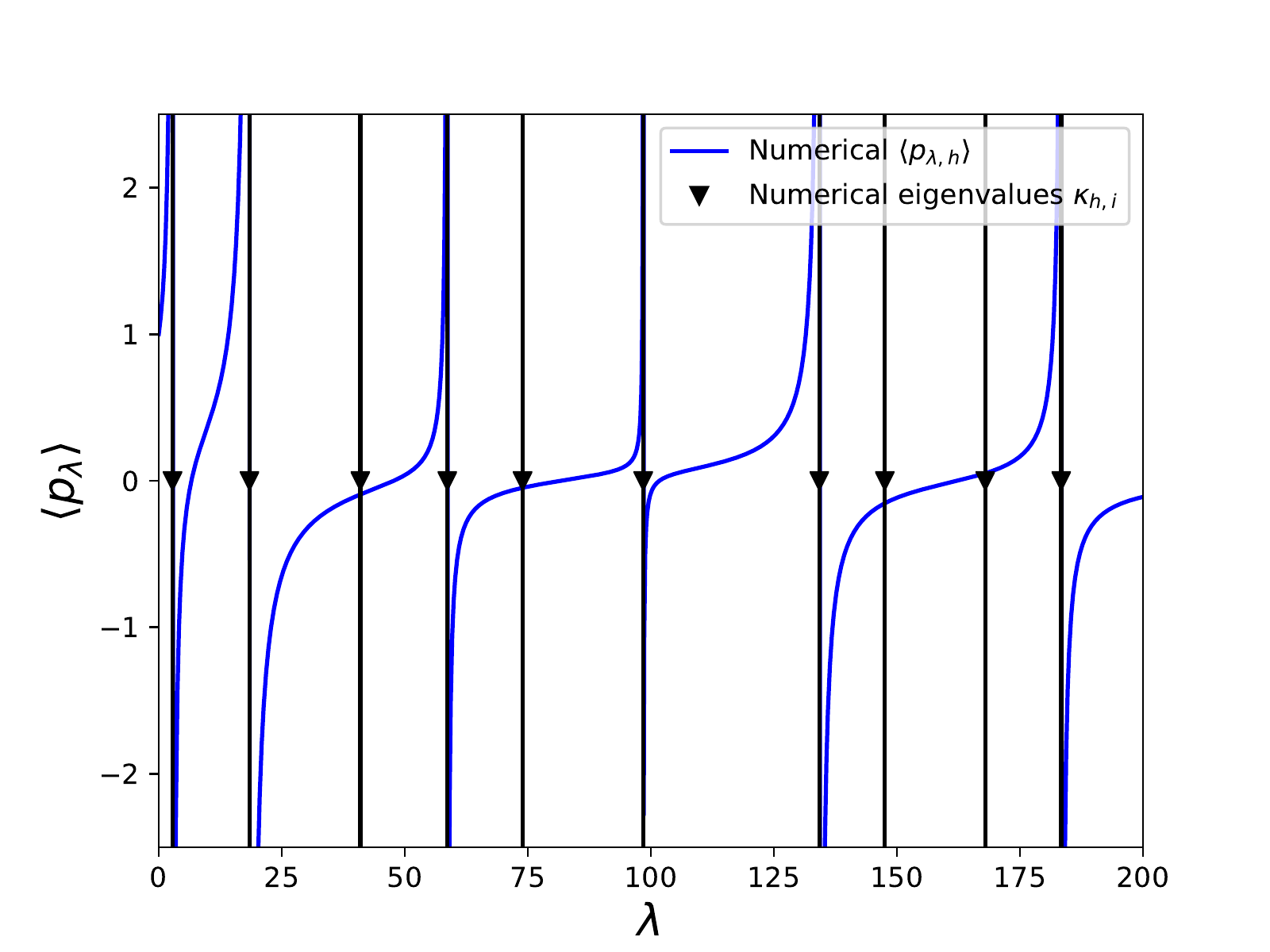}
    (b)
    \end{minipage}
    \caption{The response $p_\lambda$ for $\lambda = 20$ (a), and the average pressure $\langle p_\lambda \rangle$ (b)
             for the polygonal cylinder shown in Figure~\ref{fig:cylinders}(b).
             The vertical lines in (b) indicate the poles.}
    \label{fig:polyresponse}
\end{figure}

We will represent the solution $p_\lambda$ of \eqref{eq:FR-nonuni} in terms of the eigenpairs of the Laplace operator
\begin{align}\label{eq:evpnonuni}
-\Delta \psi & = \kappa \psi \quad\, \text{ in } \Omega, \notag\\
\psi & = 0 \quad\,\,\,\, \text{ on } \Gamma_D, \\
\nabla \psi \cdot \nu & = 0 \quad\,\,\,\, \text{ on } \Gamma_N.\notag
\end{align}
As before, the spectrum $0 < \kappa_1 < \kappa_2 \le \cdots \le \kappa_j \to \infty$ is discrete, and the eigenfunctions $\psi_i$ form a Hilbert basis in $L^2(\Omega)$, and we assume that they are orthonormalized by $\int_\Omega \psi_i \psi_j\, dx = \delta_{ij}$.
Writing $p_\lambda = 1 + \sum_{i=1}^\infty \beta_i \psi_i$ and substituting into \eqref{eq:FR-nonuni} one gets
\begin{align}
    \label{eq:p-nonuni}
    p_\lambda(x)= 1 + |\Omega| \sum_{i=1}^\infty \frac{\lambda}{\kappa_i-\lambda} \langle \psi_i\rangle \psi_i(x),
    \quad \langle \psi_i\rangle = \frac{1}{|\Omega|} \int_\Omega \psi_i \, dx.
\end{align}
The mean value of $p_\lambda$ in $\Omega$ is
\begin{align}
    \label{eq:average-p}
    \langle p_\lambda \rangle &= \frac{1}{|\Omega|} \int_\Omega p_\lambda \,dx 
     = 1 + |\Omega| \sum_{i=1}^\infty \frac{\lambda}{\kappa_i-\lambda} \langle \psi_i\rangle^2 \nonumber \\
    & = 1-|\Omega|\sum_{i=1}^\infty \langle \psi_i\rangle^2
    + |\Omega|\sum_{i=1}^\infty \frac{\lambda}{\kappa_i-\lambda} \langle \psi_i\rangle^2.
\end{align}
The pressure $p_\lambda$ in a polygonal cylinder, and its mean-value $\langle p_\lambda \rangle$ with respect to the space variable as a function of $\lambda$ is shown in Figure~\ref{fig:polyresponse}.

Let us now average \eqref{eq:average-p} over $(\lambda_{\text{min}}, \lambda_{\text{max}})$ to get the objective function:
\begin{align}\label{eq:objectivep}
    \Psi &= \frac{1}{\lambda_{\text{max}} - \lambda_{\text{min}}}
    \text{p.v.}\int_{\lambda_{\text{min}}}^{\lambda_{\text{max}}}
    \langle p_\lambda \rangle \, d\lambda \notag \\
    &= 
    1 
    + |\Omega| \sum_{i=1}^\infty \left[\frac{\kappa_i}{\lambda_{\text{max}} - \lambda_{\text{min}}} \log \left|\frac{\kappa_i-\lambda_{\text{min}}}{\lambda_{\text{max}}-\kappa_i}\right| -1 \right]\langle \psi_i\rangle^2.
\end{align}
As we have seen above, for the case of a uniform cylinder,
the right hand side of \eqref{eq:objectivep} sums up to~\eqref{eq:psi-exact}.

%\clearpage

%%%%%%%%%%%%%%%%%%%%%%%%%%%%%%%%%%%%%%%%%%%%%%%%%%%

\subsection{Numerical computation of the average pressure response}
\label{sec:num-psi}

In order to compute the objective function $\Psi$ \eqref{eq:objectivep} in the case of a non-uniform cylinder,
we compute the eigenpairs $(\kappa_j, \psi_j)$ of \eqref{eq:evpnonuni} using the first order Lagrange finite
elements.

The variational formulation for \eqref{eq:evpnonuni} reads: Find $\kappa \in \mathbf R$ and $\psi \in H^1(\Omega)\setminus \{0\}$, $\psi=0$ on $\Gamma_D$, such that
\begin{align}
\label{eq:var-form}
    \int_\Omega \nabla \psi \cdot \nabla v \, dx = \kappa \int_\Omega \psi\, v\, dx, 
\end{align}
for any $v\in H^1(\Omega)$, $v=0$ on $\Gamma_D$. 
For a triangulation mesh $\mathcal T_h$ of $\Omega$, we consider Lagrange triangular finite elements of order $1$ as a basis for the finite-dimensional subspace
\begin{align}
    V_{0h} =\Big\{v \in C(\overline \Omega) :  v\big|_{K} \in \mathbb P_1 \,\, \text{for all} \,\, K \in \mathcal T_h, \,\, v=0 \,\, \text{on}\,\, \Gamma_D \Big\}.
\end{align}
The internal approximation for the variational formulation \eqref{eq:var-form} is
\begin{align}
\label{eq:var-form-h}
    \int_\Omega \nabla \psi_h \cdot \nabla v_h \, dx = \kappa_h \int_\Omega \psi_h\, v_h\, dx,
\end{align}
for all $v_h \in V_{0h}$.
The eigenvalues of \eqref{eq:var-form-h} form a finite increasing sequence 
\begin{align*}
0 & < \kappa_{h,1} \le \kappa_{h,2} \le \cdots \le \kappa_{h,n_{dl}}, \quad \text{with}\,\, n_{dl}={\rm dim} \, V_{0h},
\end{align*}
and there exists a basis in $V_{0h}$ consisting of corresponding eigenfunctions which is orthonormal in $L^2(\Omega)$. A proof of this statement can be found in~\cite[Ch. 7.4]{allaire2007numerical}.

We look for a solution of \eqref{eq:var-form-h} in the form $\psi_h(x)=\sum_{i=1}^{n_{dl}} U_i^h \phi_i(x)$, where $(\phi_i)_{1\le i\le n_{dl}}$ is the basis in $V_{0h}$.
Introducing the mass matrix $\mathcal M_h$ and the stiffness matrix $\mathcal K_h$,
\begin{align}
    (\mathcal M_h)_{ij} = \int_\Omega \phi_i\, \phi_j\, dx, \quad
    (\mathcal K_h)_{ij} = \int_\Omega \nabla \phi_i\cdot \nabla \phi_j\, dx, \quad 1\le i,j\le n_{dl},
\end{align}
we get the following discrete finite-dimensional spectral matrix problem:
\begin{align}\label{eq:discreteevp}
    \mathcal K_h \psi_h = \kappa_h \mathcal M_h \psi_h.
\end{align}
The matrices $\mathcal M_h$ and $\mathcal K_h$ are symmetric and positive definite. 

The error estimate for the eigenvalues corresponding to eigenfunctions in $H^2(\Omega)$, which is for instance the case if $\Omega$ is convex in $\mathbf{R}^2$, is
%, which for instance is the case for polygonal domains in $\mathbf{R}^2$, is
\begin{align*}
|\kappa_i - \kappa_{h,i}| \le C_i h^2,
\end{align*}
where $C_i$ does not depend on $h = \max \{\mathrm{diam}(K) : K \in \mathcal{T}_h \}$,
but does depend on the number of the eigenvalue, that is why is it important to take a sufficiently fine mesh to get a good approximation for $\kappa_i$ with large $i$ (see e.g.~\cite{allaire2007numerical}).
More precisely, if $\kappa_i$ is an eigenvalue with eigenfunctions in $H^{k+1}(\Omega)$, $\Omega \subset \mathbf{R}^n$, and $2(k+1) > n$, then
$|\kappa_i - \kappa_{h,i}| \le C_i h^{2k}$.

In the numerical method, we truncate the series in \eqref{eq:objectivep} at $i = N$,
\begin{align}\label{eq:objectivep-h}
    \Psi_h &= 
    1 + |\Omega| \sum_{i=1}^N \left[\frac{\kappa_{h,i}}{\lambda_{\text{max}} - \lambda_{\text{min}}} \log \left|\frac{\kappa_{h,i}-\lambda_{\text{min}}}{\lambda_{\text{max}}-\kappa_{h,i}}\right| -1 \right]\langle \psi_{h,i}\rangle^2,
\end{align}
where $N$ is chosen such that the sum ranges over the eigenvalues up to at least $10\lambda_{\max}$,
and the number of degrees of freedom $\mathrm{dim}\,V_{0h}$ is at least $10$ times greater than the greatest eigenvalue $\kappa_{h,N}$ used in the computation.
This ensures that the eigenvalues $\kappa_{h,i}$ and the eigenfunctions $\psi_{h,i}$
of the discrete eigenvalue problem~\eqref{eq:discreteevp}
are correct approximations to the exact eigenvalues $\kappa_{h,i}$ and exact eigenfunctions
$\psi_{h,i}$ of~\eqref{eq:var-form}.

\begin{figure}[htp]
    \centering
    \begin{minipage}{.49\textwidth}
    \centering
    \includegraphics[width=.99\linewidth]{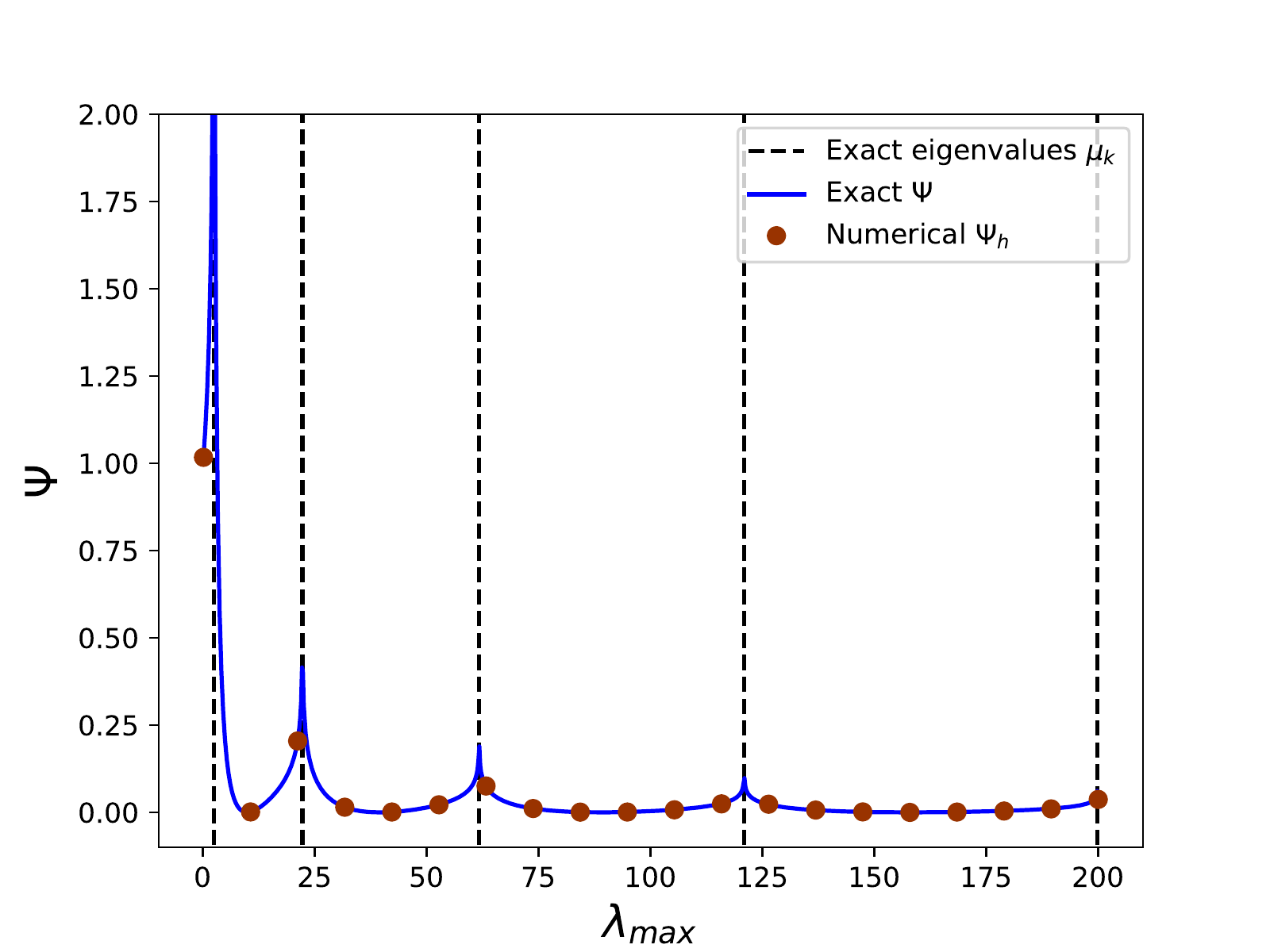}
    (a)
    \end{minipage}
    \begin{minipage}{.49\textwidth}
    \centering
    \includegraphics[width=.99\linewidth]{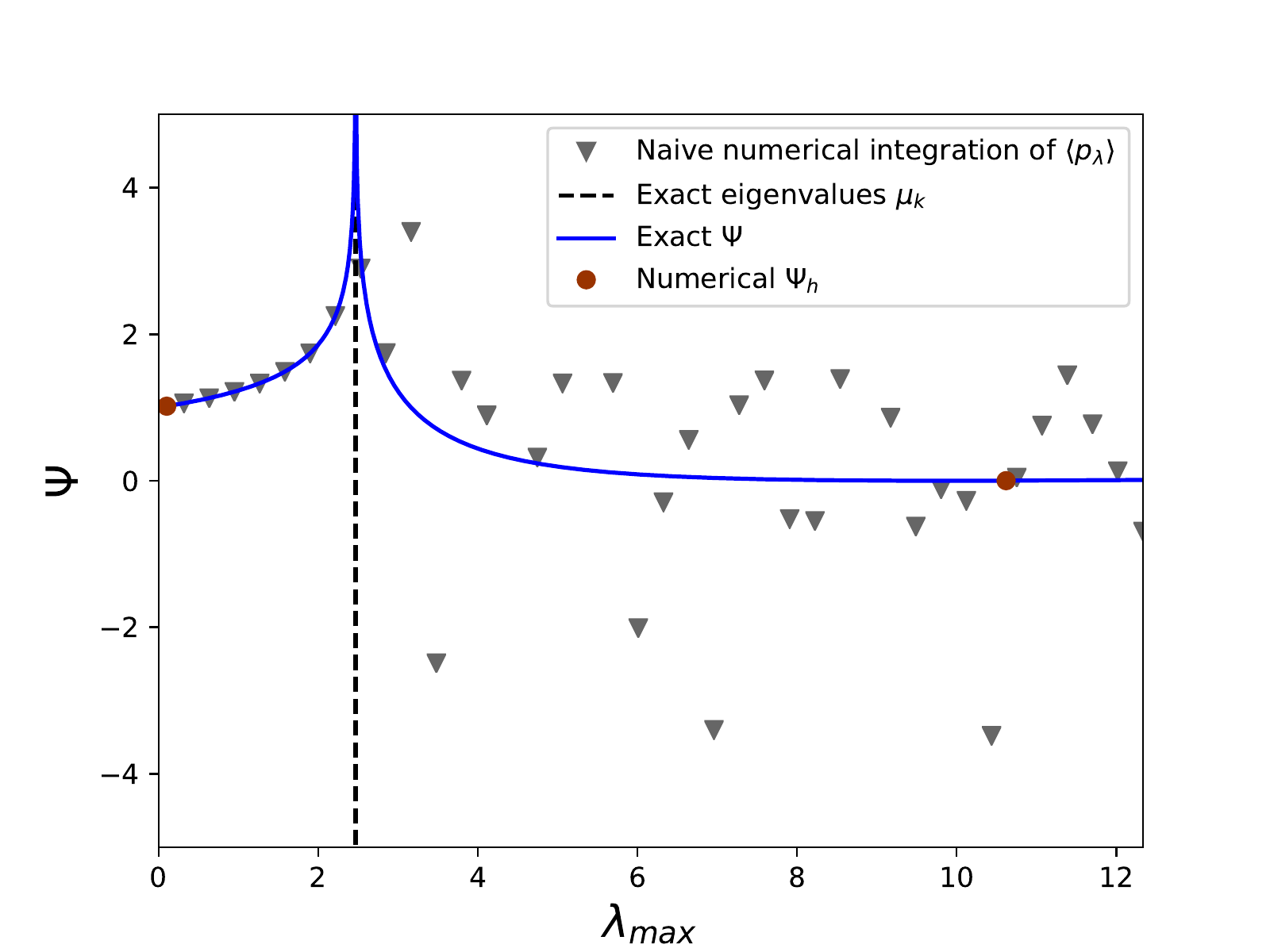}
    (b)
    \end{minipage}
    \caption{Objective function in a uniform cylinder
             for intervals $(0,\lambda_{\max})$ with $\lambda_{\max}$ on the horizontal axis.}
    \label{fig:uninum}
\end{figure}

In order to evaluate the accuracy of the method, in the case of uniform cylinders, we can compare the exact objective function \eqref{eq:psi-exact} (the blue curve) with its numerical approximation \eqref{eq:objectivep-h} (the dots). The result is presented in Figure~\ref{fig:uninum}(a). The peaks of the objective function are located at the eigenvalues $\mu_k$ since $p_\lambda$ has poles at these points. The graph is valid for a uniform cylinder of arbitrary radius, because the pressure does not depend on the transverse variable.
It is important to note that numerical integration by means of the trapezoidal rule of the exact response $\langle p_\lambda \rangle$ given by~\eqref{eq:avgp} with respect to $\lambda$ does not give a good approximation for $\Psi$. In Figure~\ref{fig:uninum}(b) one can see that the numerical integration fails after the first eigenvalue. 
The reason for this is the singular behavior of $p_\lambda$ near the eigenvalues $\mu_k$.

\begin{figure}[htp]
    \centering
    \begin{minipage}{.49\textwidth}
    \centering
    \includegraphics[width=.99\linewidth]{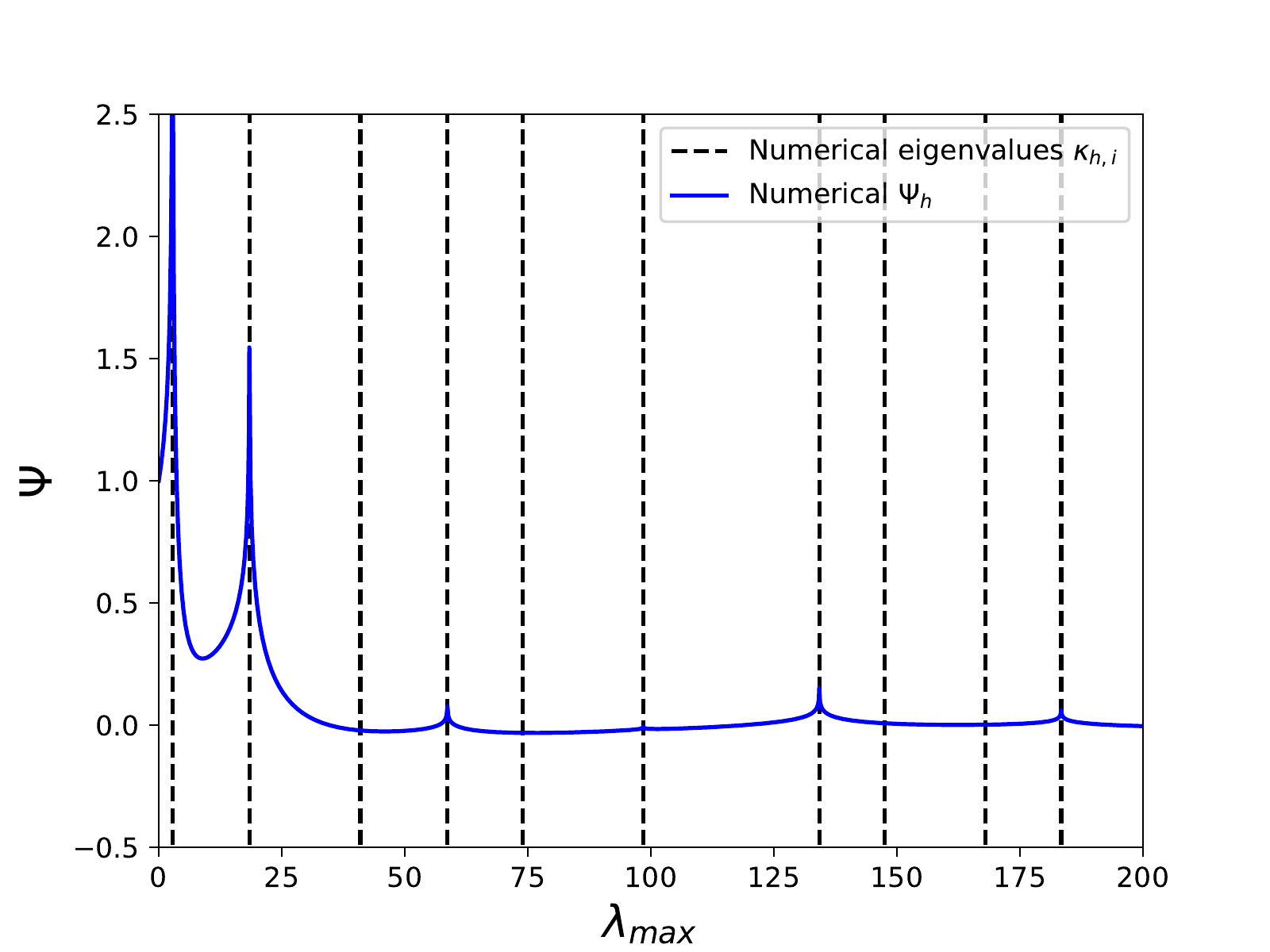}
    (a)
    \end{minipage}
    \begin{minipage}{.49\textwidth}
    \centering
    \includegraphics[width=.99\linewidth]{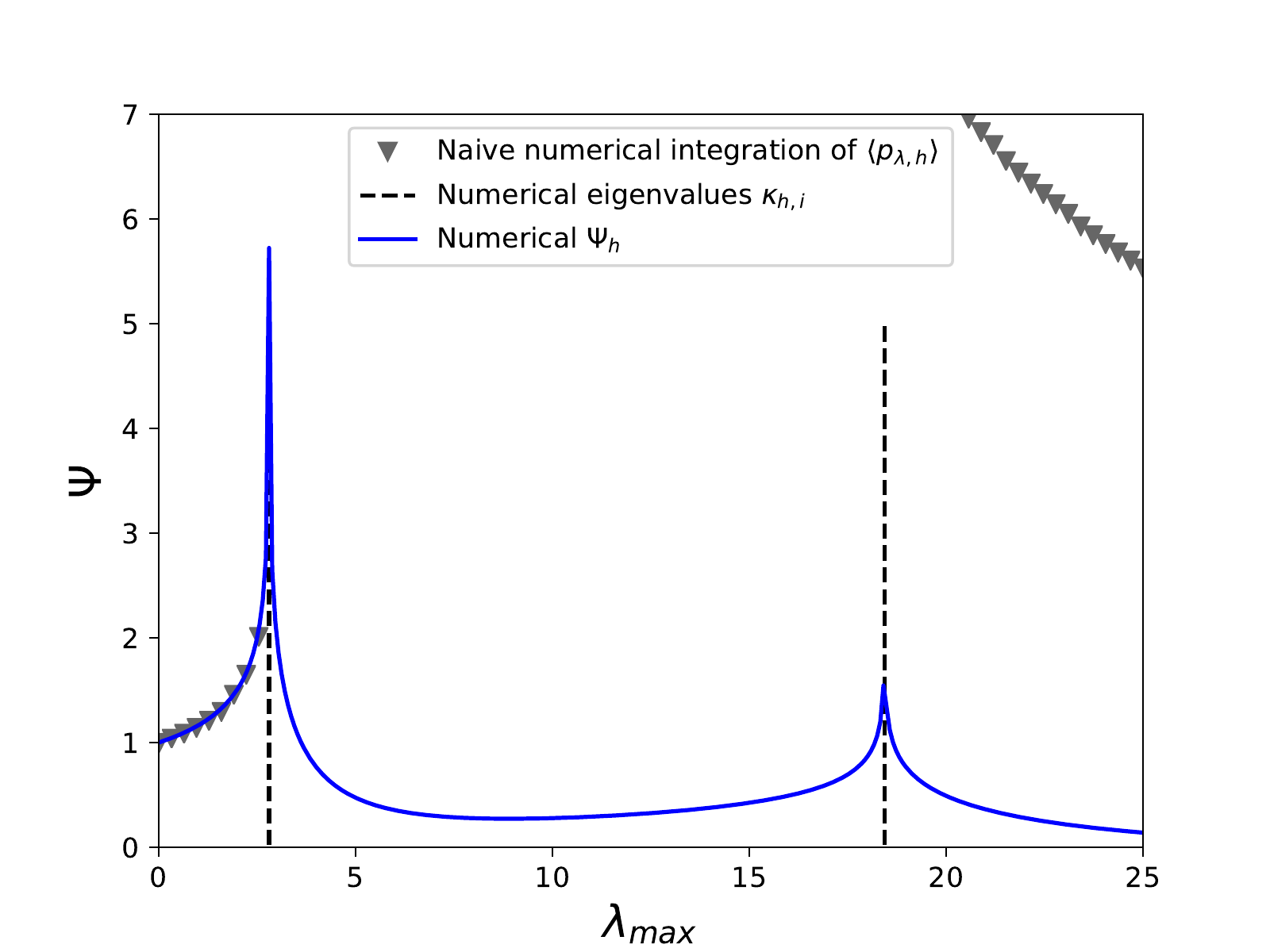}
    (b)
    \end{minipage}
    \caption{The objective function in a non-uniform cylinder in Figure~\ref{fig:cylinders}(b) for intervals $(0,\lambda_{\max})$ with $\lambda_{\max}$ on the horizontal axis.}
    \label{fig:polynum}
\end{figure}

In the case of non-uniform cylinders we do not have any explicit formulas any more, so we investigate numerically the rate of convergence for the approximation of $\Psi$ by $\Psi_h$. For the sake of completeness, we present the convergence rate for both uniform and non-uniform cylinders.

\begin{figure}[htp]
    \centering
    \begin{minipage}{.49\textwidth}
    \centering
    \includegraphics[width=\linewidth]{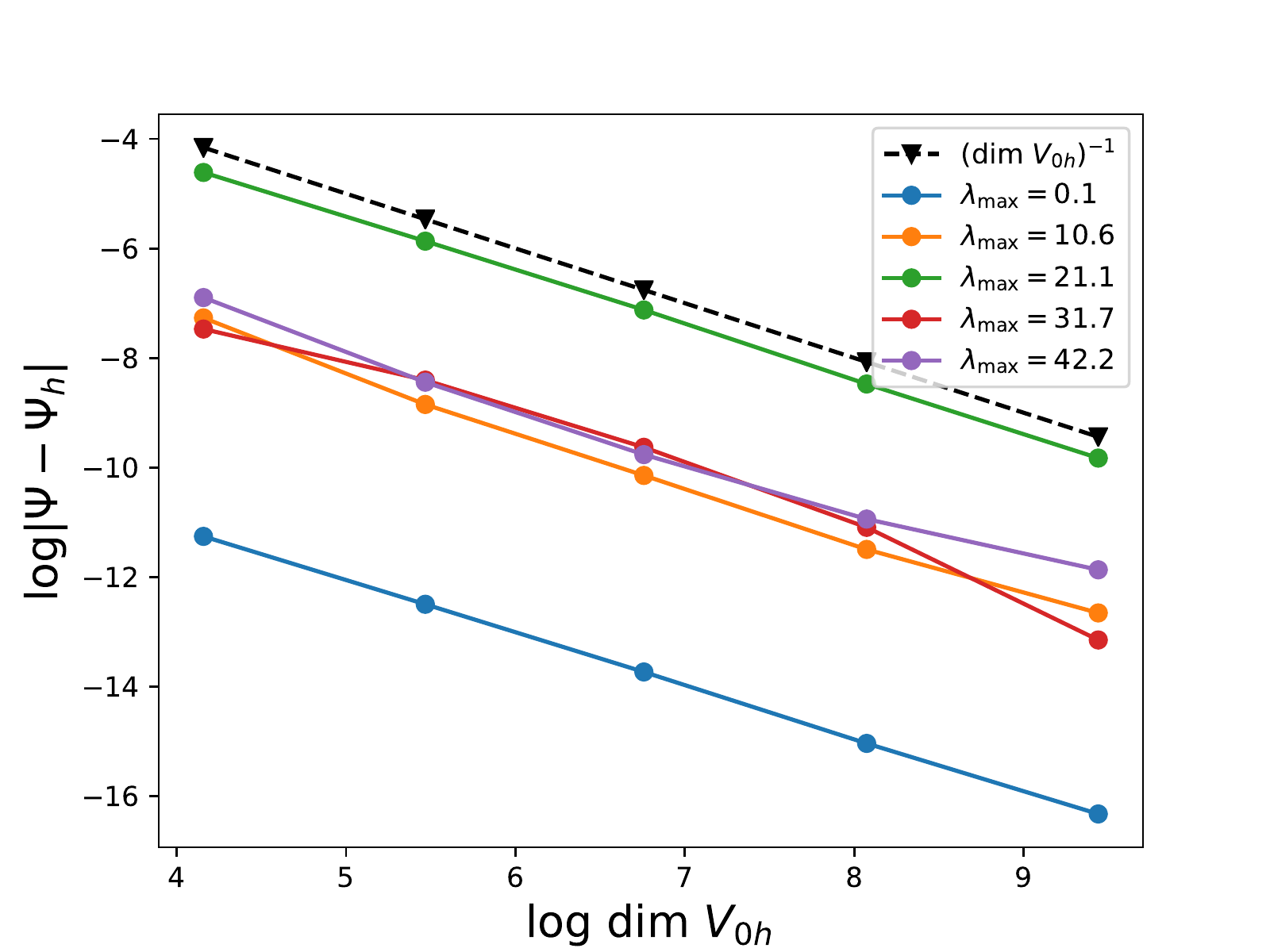}
    (a)
    \end{minipage}
    \begin{minipage}{.49\textwidth}
    \centering
    \includegraphics[width=\linewidth]{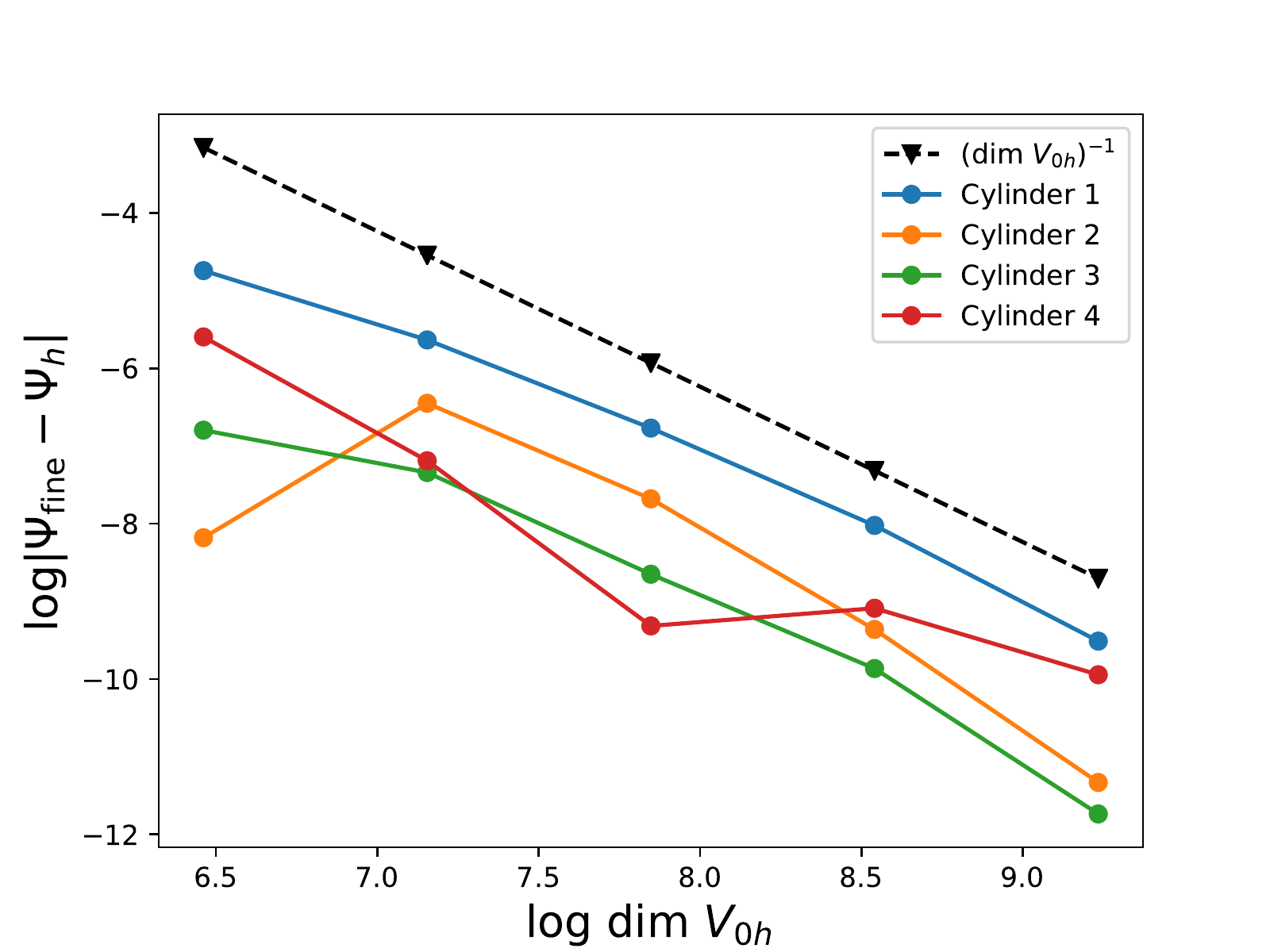}
    (b)
    \end{minipage}
    \caption{Rate of convergence of the finite element approximation of the objective function $\Psi$. (a) The absolute error for a uniform cylinder. (b) The estimated rate of convergence 
    for four samples of non-uniform cylinders, and fixed $\lambda_{\max} = 60$.}
    \label{fig:femconvergence}
\end{figure}

In Figure~\ref{fig:femconvergence}(a), one can see a clear quadratic decay of $|\Psi-\Psi_h|$ with respect to mesh size $h$ for uniform cylinders (convex).
The objective function $\Psi_h$ \eqref{eq:objectivep-h} is computed as the average over $(0,\lambda_{\max})$ for several $\lambda_{\max}$ and for uniform mesh refinements.
The quadratic decay of the error with respect to the mesh size $h$ is expected for first order polynomial approximations of a smooth function in $L^2(\Omega)$.
In $\mathbf{R}^2$, the number of degrees of freedom
$\mathrm{dim}\, V_{0h}$ grows as $h^{-2}$ for uniform mesh refinement,
which suggests an expected rate of decay $(\mathrm{dim}\,V_{0h})^{-1}$
for non-degenerate uniform mesh refinement.

In Figure~\ref{fig:femconvergence}(b), we present the rate of convergence while refining the mesh for several cylinders with polygonal boundary.
We observe a subquadratic convergence rate with respect to the mesh size.

%%%%%%%%%%%%%%%%%%%%%%%%%%%%%%%%%%%%%%%%%%%%%%%%%%%%%%%%%%%%

\subsection{Shape derivative of the average pressure response}
\label{sec:shape}

In this section we compute the derivative $\Psi'$ of the objective function
\begin{align}\label{eq:objj}
\Psi & = \frac{1}{\lambda_\mathrm{max}-\lambda_\mathrm{min}}
 \int_{\lambda_{\mathrm{min}}}^{\lambda_{\mathrm{max}}}
\langle p_\lambda \rangle \,d\lambda,
\end{align}
with respect to certain variations of the convex cylindrical domains $\Omega$ in $\mathbf{R}^2$.
The purpose of this is twofold.
One, it ascertains that the accuracy of our trained model on Lipschitz domains
that are close in a precise sense to certain polygonal domains in our evaluation sets.
Two, it enables for some boosting of the training sets that are otherwise somewhat
costly to generate by the method we have chosen.

For a vector field $V$ and a parameter $t$, we introduce the bi-Lipschitz transformation
$T(x) = x + t V(x)$. We denote by $\Omega_t$ the image of $\Omega$ under $T$.
Let $\Psi_t$ be the value of~\eqref{eq:objj} for the domain $\Omega_t$, $\Psi = \Psi_0$.
With $\Psi'_0$, the derivative of $\Psi$ at $t = 0$, $\Psi$
is linearized as
\begin{align*}
\Psi_t & = \Psi_0 + t \Psi'_0 + o(t),
\end{align*}
as $t$ tends to zero.

In this section we employ standard techniques of domain variations in the theory of elliptic equations.
We refer to \cite{necas1967methodes}, \cite{komkov1986design}, and \cite{grisvard} for expositions.

\begin{lemma}\label{lm:shape}
Let $V \in W^{1,\infty}(\Omega)$ be a solenoidal vector field on convex $\Omega$.
Suppose that the interval $[\lambda_{\min}, \lambda_{\max}]$ does not contain any eigenvalue $\kappa_i$ for which $\langle \psi_i \rangle \neq 0$.
Then the shape derivative of the objective function \eqref{eq:objj} is given by
\begin{align*}
\Psi'
& =
\lim_{t \to 0} \frac{\Psi_t - \Psi_{0}}{t}
=
%\sum_{i, j = 1}^\infty \sum_{k,l} c_{i,j} \langle \partial_k \psi_i \partial_l \psi_j \partial_k V_l \rangle
%\langle \psi_i \rangle
%\langle \psi_j \rangle \\
%& = 
\sum_{i, j = 1}^\infty c_{i,j} \langle \nabla V \nabla \psi_i \cdot \nabla \psi_j \rangle
\langle \psi_i \rangle
\langle \psi_j \rangle,
\end{align*}
where for $\kappa_i = \kappa_j$,
\begin{align*}
c_{i,j} & = \frac{2|\Omega|^2}{\lambda_{\max}-\lambda_{\min}}
\left[
\log \left| \frac{\lambda_{\max} - \kappa_i}{\kappa_i - \lambda_{\min}} \right|
- \frac{\kappa_i}{\lambda_{\max} - \kappa_i}
- \frac{\kappa_i}{\kappa_i - \lambda_{\min}}
\right],
\end{align*}
and for $\kappa_i \neq \kappa_j$,
\begin{align*}
c_{i,j} & =
\frac{2|\Omega|^2}{\lambda_{\max}-\lambda_{\min}} \left[
\frac{\kappa_i}{\kappa_i - \kappa_j}
\log \left| \frac{\lambda_{\max} - \kappa_i}{\kappa_i - \lambda_{\min}} \right|
-
\frac{\kappa_j}{\kappa_i - \kappa_j}
\log \left| \frac{\lambda_{\max} - \kappa_j}{\kappa_j - \lambda_{\min}} \right|
\right].
\end{align*}
\end{lemma}
\begin{proof}
One notes that by elliptic regularity $p_\lambda \in H^2(\Omega)$.
Denote by $\dot p_\lambda \in H^1(\Omega, \Gamma_D)$ the material derivative of $p_\lambda$:
$\dot p_\lambda = p_\lambda' + \nabla p_\lambda \cdot V$,
where $p_\lambda' \in H^1(\Omega)$ denotes the shape derivative of the response with respect to $V$. %The function $p_\lambda=p_\lambda(t, x_t)$ depends on $t$ as a solution of the corresponding boundary-value problem in $\Omega_t$ and it is evaluated at $x_t=x+tV$.
%The derivatives are defined by
%\begin{align*}
%\lim \limits_{t\to 0} \left \| \dot p_\lambda -  \frac{p_\lambda(t, x_t)-p_\lambda(x)}{t} \right\|_{H^1(\Omega)} & = 0,\\
%\lim \limits_{t\to 0} \left\| p_\lambda' - 
%    \frac{p_\lambda(t,x)-p_\lambda(x)}{t} \right\|_{H^1(\Omega)} & = 0.
%\end{align*}
By the regularity of $V$, there exist a Sobolev extension, and thereby 
the shape derivatives with respect to $V$ of the response 
and the associated linear and bilinear forms exist in the sense of 
Fr{\'e}chet with respect to the parameter $t$.

A direct computation of the Gateaux derivative gives
\begin{align*}
\Psi'
& =
\frac{1}{\lambda_{\max} - \lambda_{\min}}\int_{\lambda_{\min}}^{\lambda_{\max}} 
\langle  p_\lambda \rangle' \,d\lambda \\
& =
\frac{1}{\lambda_{\max} - \lambda_{\min}}\int_{\lambda_{\min}}^{\lambda_{\max}} (
\langle \dot p_\lambda \rangle + \langle p_\lambda \mathrm{div}V \rangle 
- \langle p_\lambda \rangle \langle \mathrm{div}V \rangle ) \,d\lambda.
\end{align*}
To compute $\langle \dot p_\lambda \rangle$, we note that $\dot p_\lambda$
is an admissible test function in the variational form of the equation for $p_\lambda$:
\begin{align*}
\int_{\Omega} \nabla (p_\lambda - 1) \cdot \nabla v \,dx
- \lambda \int_\Omega (p_\lambda - 1) v \,dx & = \lambda \int_\Omega v \,dx.
\end{align*}
Therefore,
\begin{align*}
\lambda \int_\Omega \dot p_\lambda \,dx
& = 
\int_{\Omega} \nabla (p_\lambda - 1) \cdot \nabla \dot p_\lambda \,dx
- \lambda \int_\Omega (p_\lambda - 1) \dot p_\lambda \,dx \\
%& = 
%- a'_V(\tilde p_\lambda, p_\lambda) + \lambda d'_V(\tilde p_\lambda, p_\lambda) \\
& = 2 \int_\Omega \nabla (\nabla p_\lambda \cdot V) \cdot p_\lambda \,dx
- \int_\Omega \mathrm{div}(|\nabla p_\lambda|^2 V) \,dx \\
& \quad - 2 \lambda \int_\Omega (\nabla p_\lambda \cdot V) p_\lambda + \lambda 
\int_\Omega \nabla p_\lambda \cdot V \,dx
+ \lambda \int_\Omega \mathrm{div}( p_\lambda(p_\lambda - 1)V ) \,dx,
\end{align*}
where one in the second step has differentiated the variational form of the equation for $p_\lambda$ and in that way eliminated $\dot p_\lambda$
by using $p_\lambda - 1$ as a test function.
Indeed,
\begin{align*}
& \int_\Omega \nabla \dot p_\lambda \cdot \nabla v \,dx
- \lambda \int_\Omega \dot p_\lambda v \,dx \\
& \quad = - \int_\Omega \nabla (\nabla p_\lambda \cdot V) \cdot \nabla v \,dx
- \int_\Omega \nabla p_\lambda \cdot (\nabla v \cdot V) \,dx 
 + \int_\Omega \mathrm{div}( (\nabla p_\lambda \cdot \nabla v)V )  \,dx \\
& \qquad + \lambda \int_\Omega (\nabla p_\lambda \cdot V)v \,dx 
 + \lambda \int_\Omega p_\lambda (\nabla v \cdot V) \,dx 
 - \lambda \int_\Omega \mathrm{div}( p_\lambda v V ) \,dx,
\end{align*}
for any $v \in H^2(\Omega) \cap H^1(\Omega, \Gamma_D)$.
After some manipulation of terms, one concludes that
\begin{align*}
\langle p_\lambda \rangle'
& = \langle \dot p_\lambda \rangle + \langle p_\lambda \mathrm{div}V \rangle 
- \langle p_\lambda \rangle \langle \mathrm{div}V \rangle \\
%& = - \langle p_\lambda \rangle \langle \mathrm{div}V \rangle
%+ \langle p_\lambda^2 \mathrm{div}V \rangle 
% - \frac{1}{\lambda} \langle |\nabla p_\lambda|^2 \mathrm{div}V \rangle 
%+ \frac{2}{\lambda} \sum_{k,l}\langle \partial_k p_\lambda \partial_l p_\lambda
%\partial_k V_l \rangle,
& = - \langle p_\lambda \rangle \langle \mathrm{div}V \rangle
+ \langle p_\lambda^2 \mathrm{div}V \rangle 
 - \frac{1}{\lambda} \langle |\nabla p_\lambda|^2 \mathrm{div}V \rangle 
+ \frac{2}{\lambda} \langle \nabla V \nabla p_\lambda \cdot \nabla p_\lambda \rangle,
\end{align*}
which for solenoidal $V$ reduces to
\begin{align*}
\langle p_\lambda \rangle' & =
 %\frac{2}{\lambda} \sum_{k,l}\langle \partial_k p_\lambda \partial_l p_\lambda
%\partial_k V_l \rangle.
\frac{2}{\lambda} \langle \nabla V \nabla p_\lambda \cdot \nabla p_\lambda \rangle.
\end{align*}
By substituting the expansion
\begin{align*}
p_\lambda & =  1 + \sum_{i = 1}^\infty \beta_i \psi_i, \qquad \beta_i = |\Omega|\frac{\lambda}{\kappa_i - \lambda}\langle \psi_i \rangle,
\end{align*}
and integrating in $\lambda$ the desired formula is obtained, by the Fubini theorem.
\end{proof}

For instance, in the case of a uniform cylinder,
\begin{align*}
p_\lambda & = \cos(\sqrt{\lambda}x_1) + \tan(\sqrt{\lambda})
\sin(\sqrt{\lambda}x_1).
\end{align*}
For solenoidal $V \in W^{1,\infty}(\Omega)$, the 
shape derivative of the averaged pressure response is
\begin{align*}
\langle p_\lambda \rangle'
& = 
\frac{2}{\lambda} \langle (\partial_1 p_\lambda)^2 
\partial_1 V_1 \rangle \\
& = 2 \langle ( \sin(\sqrt{\lambda}x_1) - \tan(\sqrt{\lambda}) \cos(\sqrt{\lambda}x_1))^2  \partial_1 V_1 \rangle,
\end{align*}
and
\begin{align*}
\Psi' 
& =
\frac{1}{\lambda_{\max} - \lambda_{\min}}
\left\langle 
\left[ \frac{2\sin(\sqrt{\lambda})\cos(\sqrt{\lambda}x_1)^2}{\sqrt{\lambda}\cos(\sqrt{\lambda})}
- \frac{2\sin(\sqrt{\lambda}x_1)\cos(\sqrt{\lambda}x_1)}{\sqrt{\lambda}}
\right]_{\lambda_{\min}}^{\lambda_{\max}}
\frac{\partial V_1}{\partial x_1}
\right \rangle \\
& \quad +
\frac{1}{\lambda_{\max} - \lambda_{\min}}
\left\langle 
\left[ \frac{\sin(\sqrt{\lambda}x_1)\cos(\sqrt{\lambda}x_1)}{\sqrt{\lambda}\cos(\sqrt{\lambda})^2}
+ \frac{x_1}{\cos(\sqrt{\lambda})^2}
\right]_{\lambda_{\min}}^{\lambda_{\max}}
\frac{\partial V_1}{\partial x_1}
\right \rangle.\\
\end{align*}

A somewhat more direct proof of Lemma~\ref{lm:shape} goes as follows.

\begin{proof}[A second proof of Lemma~\ref{lm:shape}]
Let $\tilde p_\lambda^t$ be such that $\tilde p_\lambda^t - 1 \in H^1(\Omega_t, T(\Gamma_D))$ and
\begin{align*}
\int_{\Omega_t} \nabla \tilde p_\lambda^t \cdot \nabla v \,dx - \lambda 
\int_{\Omega_t} \tilde p_\lambda^t v \,dx & = 0,
\end{align*}
for all $v \in H^1(\Omega_t, T(\Gamma_D))$, supposing that $t$ is small enough.
Then by the Lipschitz transform of the Sobolev space, using that $T$ and $T^{-1}$ are Lipschitz, 
\begin{align*}
p^t_\lambda = \tilde p^t_\lambda \circ T   
\end{align*}
is such that $p^t_\lambda - 1 \in H^1(\Omega, \Gamma_D)$
and it is the solution to
\begin{align}\label{eq:pullback}
\int_\Omega (\nabla T^{-T}\nabla p_\lambda^t \cdot \nabla T^{-T} \nabla v) |\mathrm{det}\nabla T|\,dx
- \lambda \int_\Omega p^t_\lambda v |\mathrm{det}\nabla T| \,dx & = 0,
\end{align}
for all $v \in H^1(\Omega, \Gamma_D)$.
Here, $\nabla T^{-T}$ denotes the transpose of the inverse of the gradient of $T$.
Using that $\psi_i$ form a Hilbert basis,
let 
\begin{align*}
p^t_\lambda & = \sum_i \gamma_i(t) \psi_i.
\end{align*}
Recall that
\begin{align*}
p_\lambda & = \sum_i \beta_i \psi_i.
\end{align*}
Using that 
\begin{align*}
\nabla T^{-T} & = 1 - t \nabla V + o(t), \\
|\mathrm{det}\nabla T| & = 1 + t \mathrm{div} V + o(t),
\end{align*}
as $t$ tends to zero,
and expanding the coefficients $\gamma_i(t)$ as 
\begin{align*}
p^t_\lambda & = \sum_i (\gamma_i^0 + \gamma_i^1 t) \psi_i + o(t),    
\end{align*}
give by equation~\eqref{eq:pullback} that $\gamma_i^0 = \beta_i$ and
\begin{align*}
p^t_\lambda & = p_\lambda + t \sum_i \gamma_i^1 \psi_i + o(t),
\end{align*}
where
\begin{align*}
\gamma_i^1 & = \sum_j \frac{\beta_j}{\kappa_i - \lambda} \left[ 
 \int_\Omega \nabla V \nabla \psi_i \cdot \nabla \psi_j \,dx 
+ \int_\Omega \nabla \psi_i \cdot \nabla V \nabla \psi_j \,dx \right.\\
& \qquad\qquad\qquad\quad \left. - \int_\Omega (\nabla \psi_i \cdot \nabla \psi_j)\mathrm{div}V \,dx 
+ \lambda\int_\Omega \psi_i \psi_j \mathrm{div}V \,dx 
\right] \\
& \quad + \frac{\lambda}{\kappa_i - \lambda} \int_\Omega \psi_i \mathrm{div}V \,dx.
\end{align*}
The shape derivative may then be computed as follows:
\begin{align*}
\Psi' & =
\frac{1}{\lambda_{\max}-\lambda_{\min}} \left( \lim_{t \to 0} \frac{\int_{\lambda_{\min}}^{\lambda_{\max}} 
\langle p_\lambda^t - p_\lambda \rangle  
\,d\lambda}{t} 
- \int_{\lambda_{\min}}^{\lambda_{\max}}   \langle p_\lambda \rangle \langle \mathrm{div}V \rangle \,d\lambda \right) \\
& = 
\frac{1}{\lambda_{\max}-\lambda_{\min}} \lim_{t \to 0} \frac{\int_{\lambda_{\min}}^{\lambda_{\max}} 
\langle p_\lambda^t - p_\lambda \rangle  
\,d\lambda}{t} 
-  \langle \mathrm{div}V \rangle \Psi \\
& = 
\frac{1}{\lambda_{\max}-\lambda_{\min}} \int_{\lambda_{\min}}^{\lambda_{\max}} 
\sum_i \gamma_i^1 \langle \psi_i \rangle  
\,d\lambda
-  \langle \mathrm{div}V \rangle \Psi.
\end{align*}
For solenoidal $V$, the computation results in
\begin{align*}
\Psi' & = 
\frac{2|\Omega|^2}{\lambda_{\max}-\lambda_{\min}} \sum_{i,j} \int_{\lambda_{\min}}^{\lambda_{\max}} 
\frac{\lambda 
}{(\kappa_i - \lambda)(\kappa_j - \lambda)}
\,d\lambda
\langle \nabla V \nabla \psi_i \cdot \nabla \psi_j \rangle 
\langle \psi_i \rangle
\langle \psi_j \rangle
.
\end{align*}
After evaluation of the integrals the desired formula is again obtained.
\end{proof}

The condition that $V$ is solenoidal in Lemma~\ref{lm:shape}
is only for presentation purpose.
The case of non-solenoidal $V \in W^{1,\infty}(\Omega)$ is covered by
both of the above proofs, except for the last step of integration in $\lambda$,
which then results in lengthier formulas.

%%%%%%%%%%%%%%%%%%%%%%%%%%%%%%%%%%%%%%%%%%%%%%%%%%%%%%%%%%%

\section{Data sets}
\label{sec:datasets}

\begin{figure}[htp!]
    \centering
    \begin{minipage}{.49\textwidth}
    \centering
    \includegraphics[width=.9\textwidth]{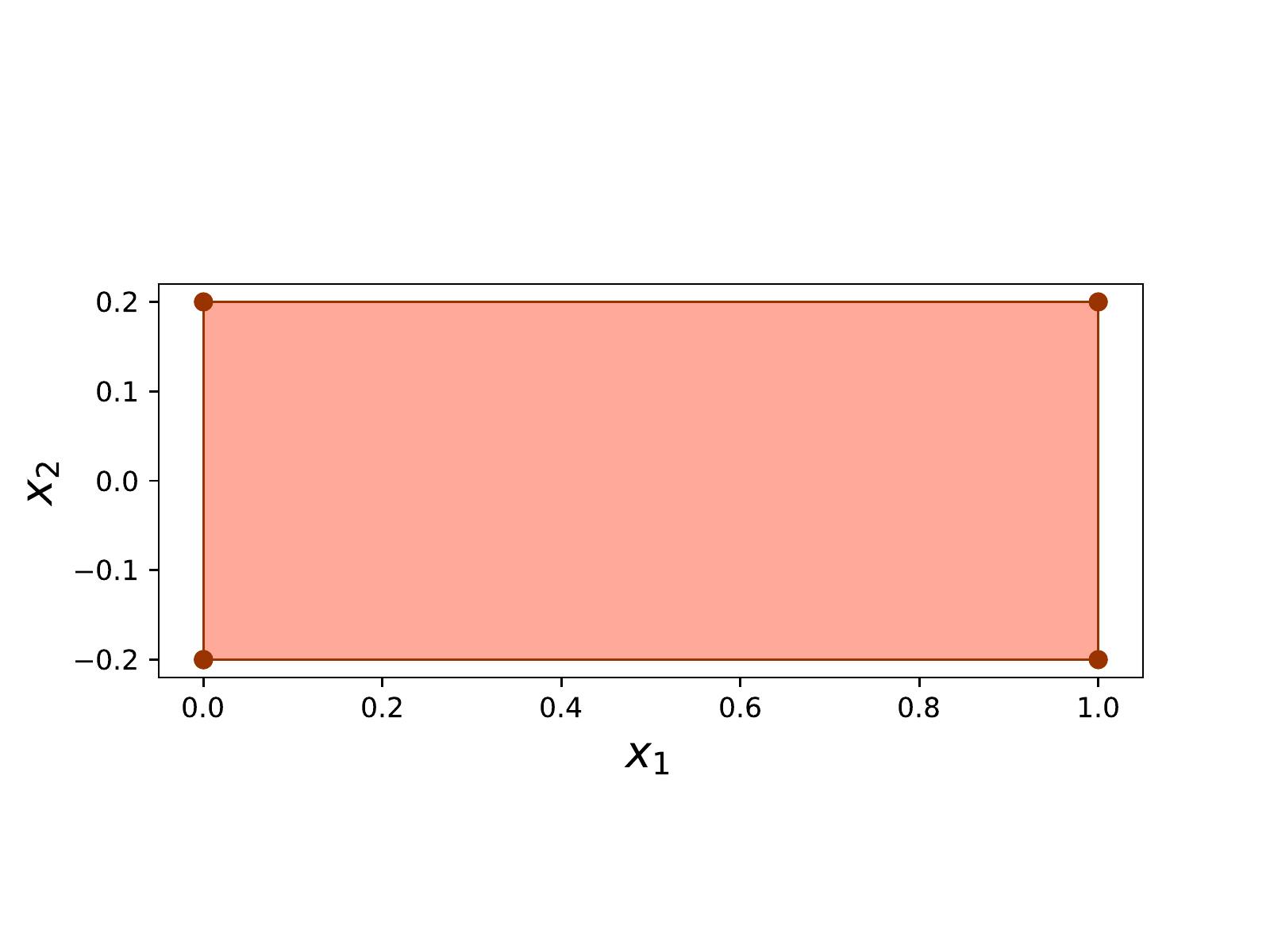}\\
    (a)
    \end{minipage}
    \begin{minipage}{.49\textwidth}
    \centering
    \includegraphics[width=.9\textwidth]{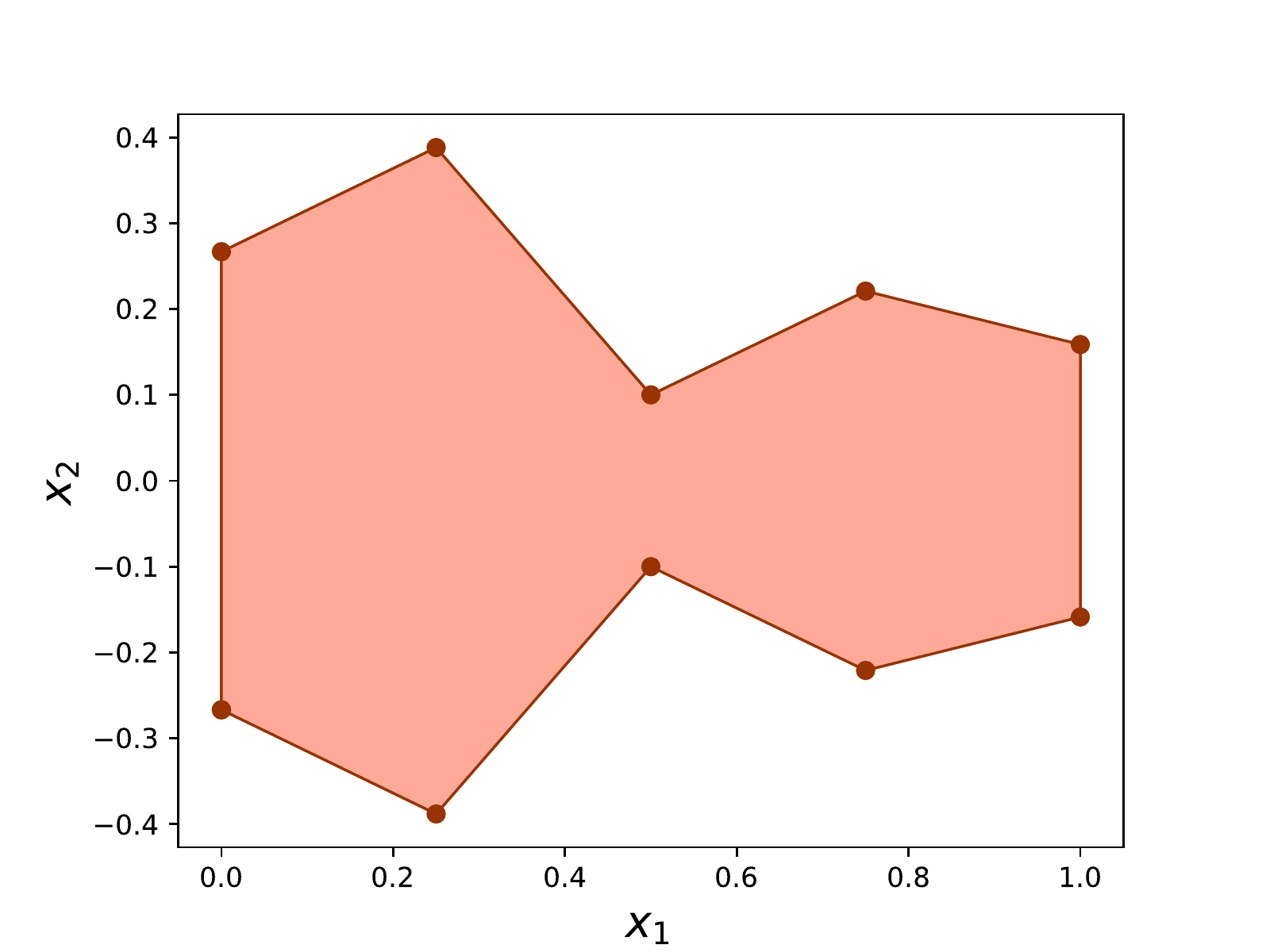}\\
    (b)
    \end{minipage}
    \caption{A uniform cylinder (a) and a non-uniform polygonal cylinder (b).}
    \label{fig:cylinders}
\end{figure}

The data sets consist of the randomly generated coordinates defining polygonal cylinders and the corresponding objective function $\Psi$.
The coordinates are generated in such a way that the radius of a cylinder varies between $0.1$ and $0.5$.
The coordinates were sampled
as independent and identically uniformly distributed random variables,
using a pseudo-random number generator.
The number of points defining the polygonal boundary might be 1 (uniform cylinder, as in Figure~\ref{fig:cylinders}(a)), 2 (cone segment), 3, 4, and 5 (as shown in Figure~\ref{fig:cylinders}(b)).
The objective function is computed, as described in the previous section, with finite elements. In total, we have about $700\,000$ data points in the main data set, which we call Random 5.
We have also generated some smaller data sets for evaluation purpose, as well as
a set of $100\,000$ data points for uniform cylinders.
The uniform cylinder set is important because it is a set for which we have very high
accuracy in the numerical value of the objective function.

For the non-uniform cylinders we have no guarantee that the error is small, as we are doing non-rigorous numerics with finite elements and floating point arithmetic without tracing or bounding the round off errors.
In the choice of mesh sizes we have employed standard indicators such as numerically observing what happens to the solution and the objective function under mesh refinement.
By means of Lemma~\ref{lm:shape}, we can guarantee for certain intervals and modulo round off errors that the error stays below a threshold $\tau$ for small enough domain perturbations of the convex
cylinders if the error on the reference is bounded by $2\tau$.
This can be exemplified with perturbations of uniform cylinders under bi-Lipschitz mappings close to the unit.
For the method of validated numerics, bounding the round off errors, we refer to~\cite{tucker2011validated}.

In Table~\ref{tab:datasets},
an overview of the data sets is provided, where we have indicated 
the size of different data subsets used for training and evaluation (test),
as well as the statistics of mean, variance, minimum, and maximum of the objective
function $\Psi_h$.

The data and the code for data generation is available on the GitHub~\cite{Pettersson2020}.

\begin{table}[hbp!]
\centering
\begin{tabular}{lrlrrrr}
\textbf{Data Set} & \textbf{Size} & \textbf{Category} & \textbf{Mean} & \textbf{Variance} & \textbf{Min} & \textbf{Max} \\
Random 5          & 200\,000        & Training    & 0.0638 & 0.00860 & -0.0918 & 0.882 \\
Random 5          & 500\,000        & Test        & 0.0636 & 0.00864 & -0.0909 & 0.991 \\
Random 5 (fine)   & 20\,000    & Test        & 0.0632 & 0.00869 & -0.0876 & 0.734  \\
Random 3          & 10\,000         & Test        & 0.0833 & 0.00494 & -0.0508 & 0.631 \\
Random 2          & 10\,000         & Test        & 0.0769 & 0.00137 & -0.0267 & 0.186 \\
Uniform           & 100\,000        & Test        & 0.0742 & 0 & 0.0742 & 0.0742\\
&
\end{tabular}
\caption{Data sets split into training and test categories.
The statistics mean, variance, min, and max of $\Psi_h$ are truncated.}
\label{tab:datasets}
\end{table}

\section{Feedforward dense neural network for approximation of average pressure}
\label{sec:dnn}

As a base model for the average pressure we will use a feedforward fully connected neural network,
with the radii of the cylinder at a discrete set of points (1, 2, 3 or 5) as input.
The base model is nonlinear, it consists of three hidden layers, each with ReLU 
activation.
We will compare this with a linear model as a point of reference.

\subsection{Structure of the neural network}

The main goal of the paper is to construct a learned algorithm which for a given polygonal cylinder outputs the corresponding average pressure. 
In this section we describe the main ideas and principles underlying the dense neural network which is used for the prediction of the average pressure level $\Psi$ over a given frequency range.
For a rigorous and at the same time concise description of the the deep neural networks construction we refer to~\cite{strang2019linear} and~\cite{montavon2012neural}.

Let us call $\Psi_\mathrm{ml}$ the function that for given cylinders outputs the objective function $\Psi_h$.
Inputs to this function are the radial coordinates of the points defining the boundary, 5 along a uniform segmentation of the interval $[0,1]$.
The output is one real number $\Psi_\mathrm{ml}$, that is we have a regression type of problem.
Assume that we have a data set containing values of $\Psi$ for $N$ polygonal cylinders.
We will train a learning function on a part of this set.
Assigning weights to the inputs, we create a function so that the error in the approximation of $\Psi$ is minimized.
Then we evaluate the performance of our function $\Psi_\mathrm{ml}$ by applying it to the unseen data and measure the accuracy of the predicted average pressures.

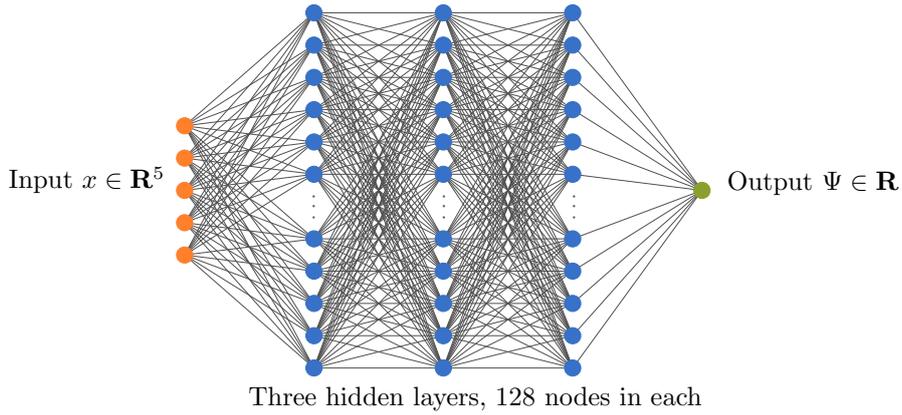
\begin{figure}[htp!]
\centering 
\def\svgwidth{.9\textwidth}
%% Creator: Inkscape 1.0 (4035a4f, 2020-05-01), www.inkscape.org
%% Accompanies image file 'nn.pdf' (pdf, eps, ps)
\begingroup%
  \makeatletter%
  \providecommand\color[2][]{%
    \errmessage{(Inkscape) Color is used for the text in Inkscape, but the package 'color.sty' is not loaded}%
    \renewcommand\color[2][]{}%
  }%
  \providecommand\transparent[1]{%
    \errmessage{(Inkscape) Transparency is used (non-zero) for the text in Inkscape, but the package 'transparent.sty' is not loaded}%
    \renewcommand\transparent[1]{}%
  }%
  \providecommand\rotatebox[2]{#2}%
  \newcommand*\fsize{\dimexpr\f@size pt\relax}%
  \newcommand*\lineheight[1]{\fontsize{\fsize}{#1\fsize}\selectfont}%
  \ifx\svgwidth\undefined%
    \setlength{\unitlength}{797.90982056bp}%
    \ifx\svgscale\undefined%
      \relax%
    \else%
      \setlength{\unitlength}{\unitlength * \real{\svgscale}}%
    \fi%
  \else%
    \setlength{\unitlength}{\svgwidth}%
  \fi%
  \global\let\svgwidth\undefined%
  \global\let\svgscale\undefined%
  \makeatother%
  \begin{picture}(1,0.47811023)%
    \lineheight{1}%
    \setlength\tabcolsep{0pt}%
    \put(0,0){\includegraphics[width=\unitlength,page=1]{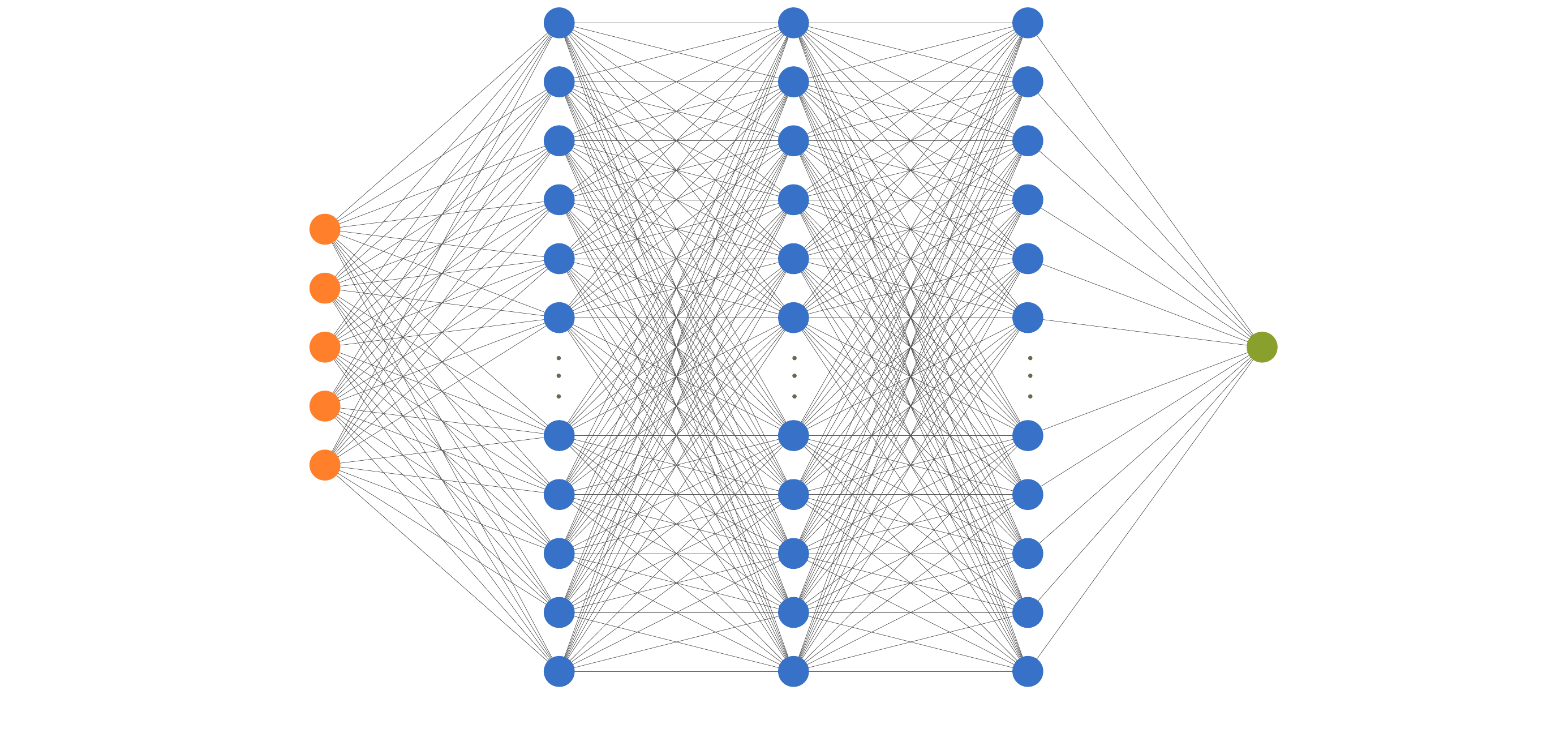}}%
    \put(0.00394524,0.25950945){\makebox(0,0)[lt]{\lineheight{1.25}\smash{\begin{tabular}[t]{l}Input $x \in \mathbf R^5$\end{tabular}}}}%
    \put(0.83410517,0.25682286){\makebox(0,0)[lt]{\lineheight{1.25}\smash{\begin{tabular}[t]{l}Output $\Psi \in \mathbf R$\end{tabular}}}}%
    \put(0.28097075,0.00696889){\makebox(0,0)[lt]{\lineheight{1.25}\smash{\begin{tabular}[t]{l}Three hidden layers, 128 nodes in each\end{tabular}}}}%
  \end{picture}%
\endgroup%

\caption{A feedforward network with three hidden layers.}
\label{fig:NN}
\end{figure}

The simplest learning function is affine, but it is usually too simple to give a good result. In Section \ref{sec:lin-vs-nonlin}, for the sake of illustration, we compare the results for linear regression and the proposed algorithm, and show that linear regression gives a poor result for nonuniform cylinders.
A widely used choice of nonlinearity is a composition of linear functions with so-called ``sigmoidal" functions (having S-shaped graph).
A smooth sigmoidal function has been a popular choice, but after that numerous numerical experiments indicated that this might not be an optimal choice.
In many examples, it has turned out that a piecewise linear function $\mathrm{ReLU}(x)=\mathrm{max}\{0,x\}$ (the positive part $x^+$ of the linear function $x$, sometimes called a rectified linear unit) performs better \cite{strang2019linear}.
Specifically, we consider a learning function $\Psi_{\mathrm{ml}}$ in the form of a composition
\begin{align}\label{eq:F}
\Psi_\mathrm{ml}(v) &= L_{M}(R(L_{M-1}(R\cdots (L_1v)))),
%\Psi_\mathrm{ml}(v) &= L_{M} \circ R \circ L_{M-1}  \circ R  \circ \cdots  \circ  L_1v,
\end{align}
where $L_k v = A_k v+b_k$ are affine functions, and  $Rx=\mathrm{ReLU}(x)$ is the nonlinear ramp function (rectifies linear unit), the activation function.
In this way the output is a recursively nested composition function of inputs: input to the first hidden layer, input from the first to the second hidden layer, $\ldots$, input from the last hidden layer to output layer. Each hidden layer in Figure~\ref{fig:NN}
contains both the linear $L_k$ and the nonlinear activation function $R$.
For our purpose seems sensible to have three hidden layers with $128$ nodes in each layer.
In this sense, we use a what could be called a deep neural network.
The elements of the matrices $A_k$ and the bias vectors $b_k$ are weights in our learning function. Note that to have $128$ nodes in the first hidden layer, the first matrix $A_1$ should have $128$ rows and $5$ columns.
The goal of the learning is to choose the weights to minimize the error over training sample,
such that it generalize well to unseen data.

\vskip 1cm

\subsection{Hyperparameters and training}
\label{sec:hyper}

The choice of hyperparameters is important for the learning of the model.
We choose the following:
\begin{itemize}
\item
\textbf{Nonlinearity:} ReLU.
    \item
    \textbf{Hidden layers:} Three hidden layers, $128$ nodes in each. 
     \item 
    \textbf{Optimizer:} ADAM.
    \item
    \textbf{Learning rate:} The step size $s_k$ in the gradient descent is scheduled to have polynomial decay from $s_0=0.001$ to $0.0001$ according to $s_k=s_0/\sqrt{k}$ in $10\,000$ steps. 
    \item
    \textbf{Loss function:} Mean squared error (MSE).
    \item
    \textbf{Validation split:} $20\%$ of the training set. 
    \item
    \textbf{Early stopping:} In order to avoid overtraining, the change of the MSE for the validation set $10^{-5}$ counts as an improvement. If we have $25$ iterations without improvement, we stop and use the weights that give the minimum MSE up to this point.   
%    \item\textbf{Minibatch size:} $32$.
    \item \textbf{Initializer:} GlorotUniform.
\end{itemize}
In Appendix A, we provide a hyperparameter grid that indicates together with the results of Section~\ref{sec:evaluation}, that the performance of the model is not that sensitive to the values of the parameters
around the chosen ones.

\section{Performance of the feedforward neural network model}
\label{sec:evaluation}

Our aim is to construct a ML algorithm to approximate $\Psi$
based on the data for $\Psi_h$ (computed with finite elements) with the same accuracy on the unseen data 
as the numerical error $\Psi - \Psi_h$.
Here, we present the measured performance on our data sets.

\begin{figure}[htp]
\begin{subfigure}{.49\textwidth}
\centering
\includegraphics[height=0.21\textheight]{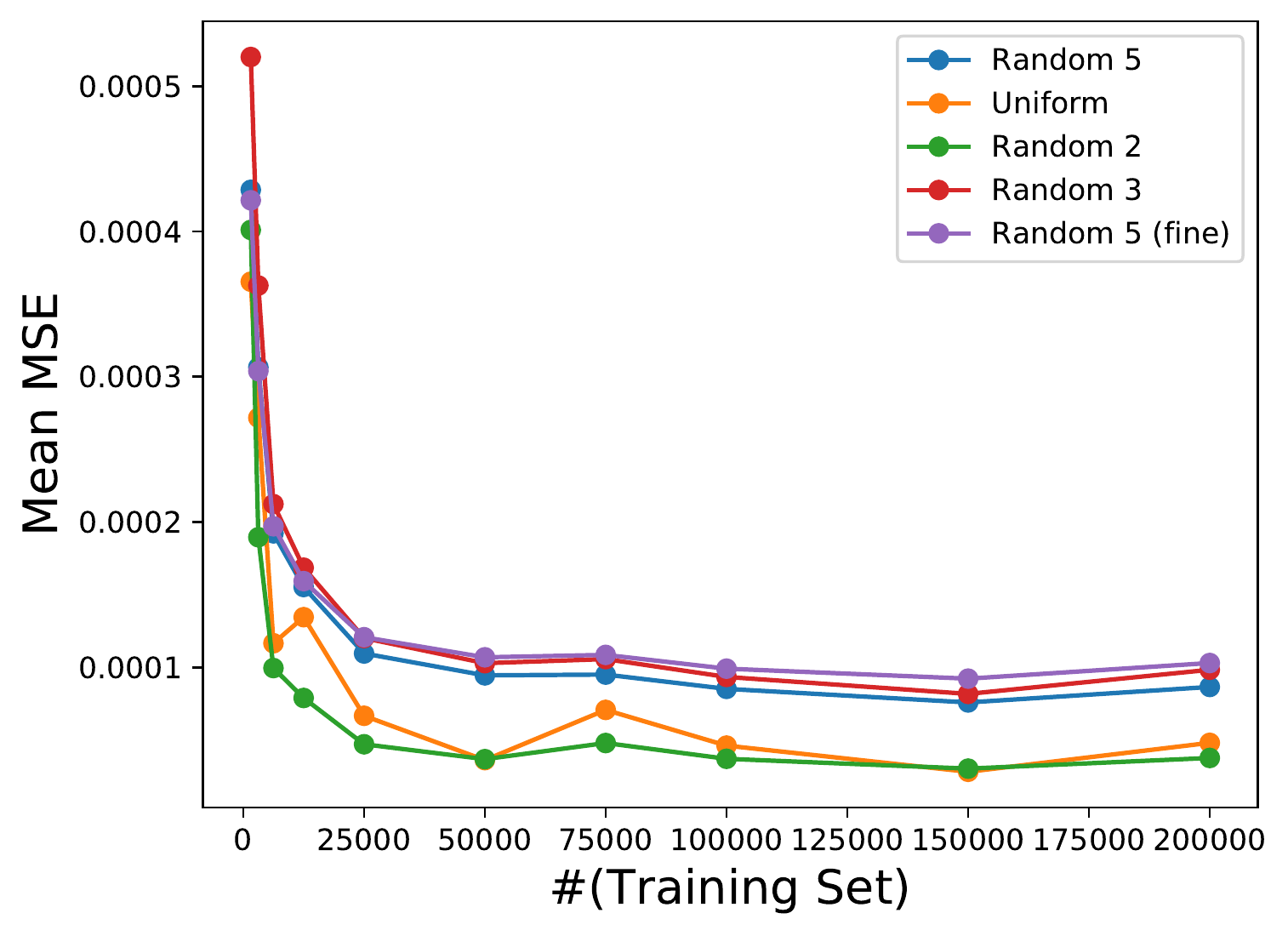}
\caption*{(a)}
\end{subfigure}
\begin{subfigure}{.49\textwidth}
\centering
\includegraphics[height=0.21\textheight]{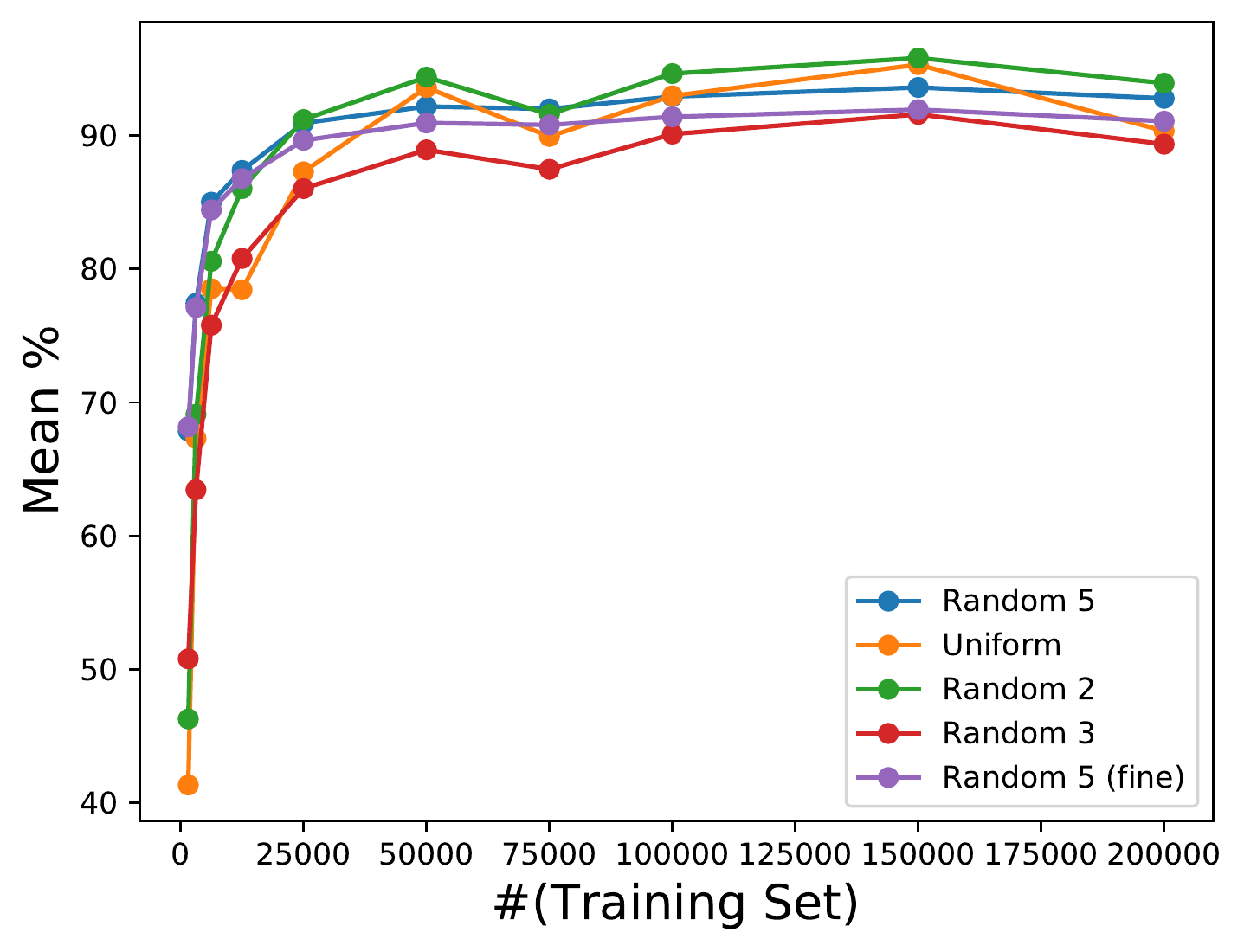}
\caption*{(b)}
\end{subfigure}
\caption{(a) Dependence of the MSE on the training set size.
(b) Percentage of the unseen data with absolute error less than $0.01$ as a function of the training set size.}
\label{fig:error-1}
\end{figure}

In Figure~\ref{fig:error-1}, we present the dependence of the error on the size of the training set. For each training set, we train the model ten times and take the mean of the mean squared error (mean MSE). The purple curve in Figure~\ref{fig:error-1}(a) shows the MSE for polygonal cylinders with five random points defining the boundary. The objective function $\Psi_h$ is computed on a mesh with density approximately three times higher than regular (referred to as ``fine").
In Figure~\ref{fig:error-1}(b),
we present the percentage of the unseen data used for the test that gives the mean absolute error less than $0.01$.
Again, here we take the mean value of the percentage after ten training sessions,
the reason being the stochastic gradient descent algorithms used which results
in some nonzero variance.

The choice of the threshold $0.01$ is based on numerical indication
of what is a bound on the error for almost all data points.
We do not guarantee this bound on the error in the numerical data.
In in spite of that, we believe it serves as an illustrative example in that similar behavior 
in the accuracy of the machine learning model on unseen data is expected
if this threshold is increased, or if the error in the data had been zero.

Numerical values for the best model are presented in Tables~\ref{tab:evamse} and \ref{tab:evaaethr} in~Section~\ref{sec:lin-vs-nonlin}. For example, for polygonal cylinders defined by five randomly generated points, over $95\%$ are predicted with mean absolute error  $0.01$ (the accuracy of the numerical data) if the training set contains $200\,000$ data points. The MSE for our model trained on $200\,000$ data points is $2.31\cdot 10^{-5}$ for uniform cylinders and $5.5\cdot 10^{-5}$ for polygonal cylinders.
Clearly, the MSE is much smaller than the variance in the data for the test set Random 5 (Table~\ref{tab:datasets}).

\begin{figure}[htp]
\begin{subfigure}{.49\textwidth}
\centering
\includegraphics[height=0.21\textheight]{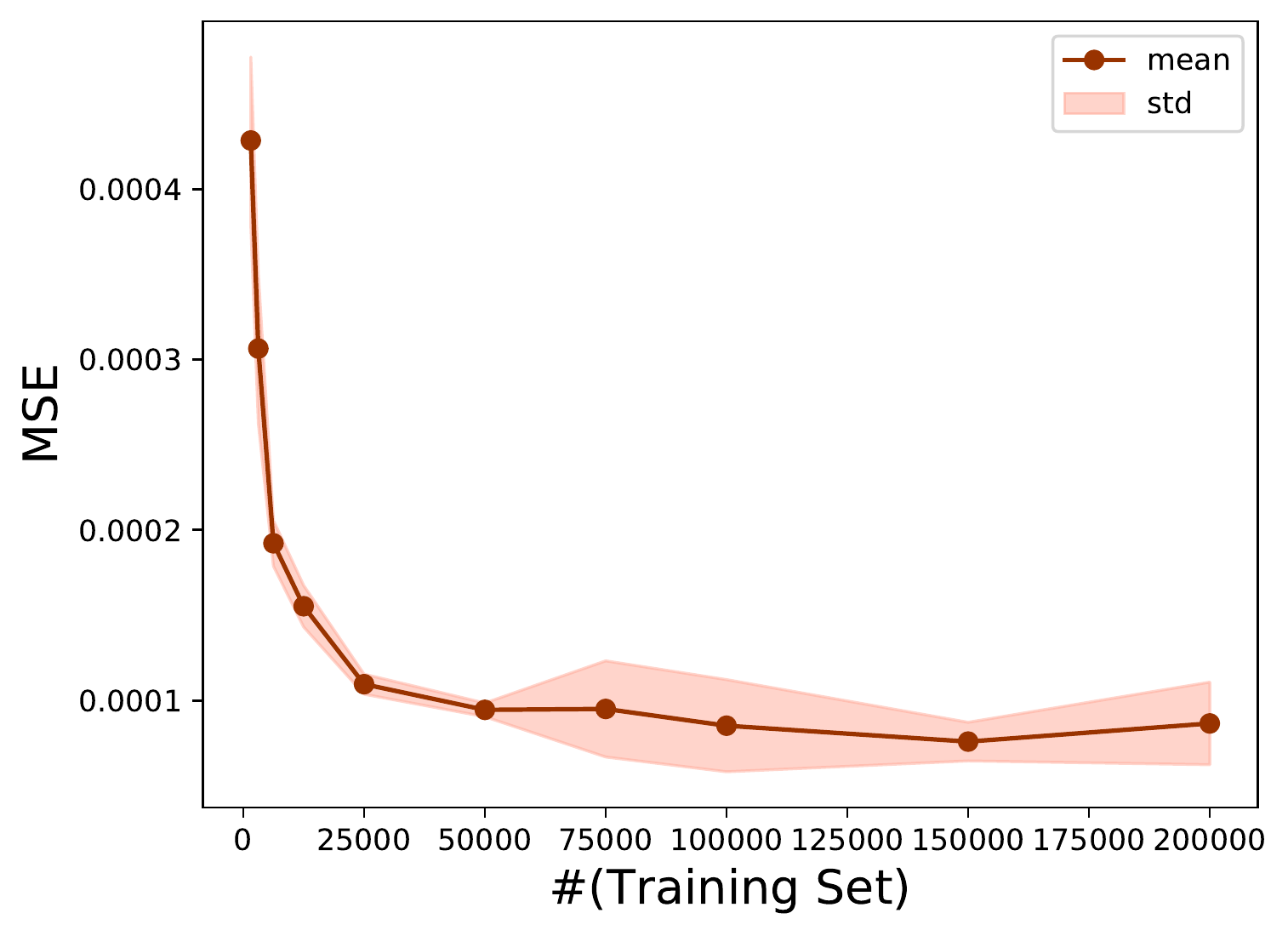}
\caption*{(a)}
\end{subfigure}
\begin{subfigure}{.49\textwidth}
\centering
\includegraphics[height=0.21\textheight]{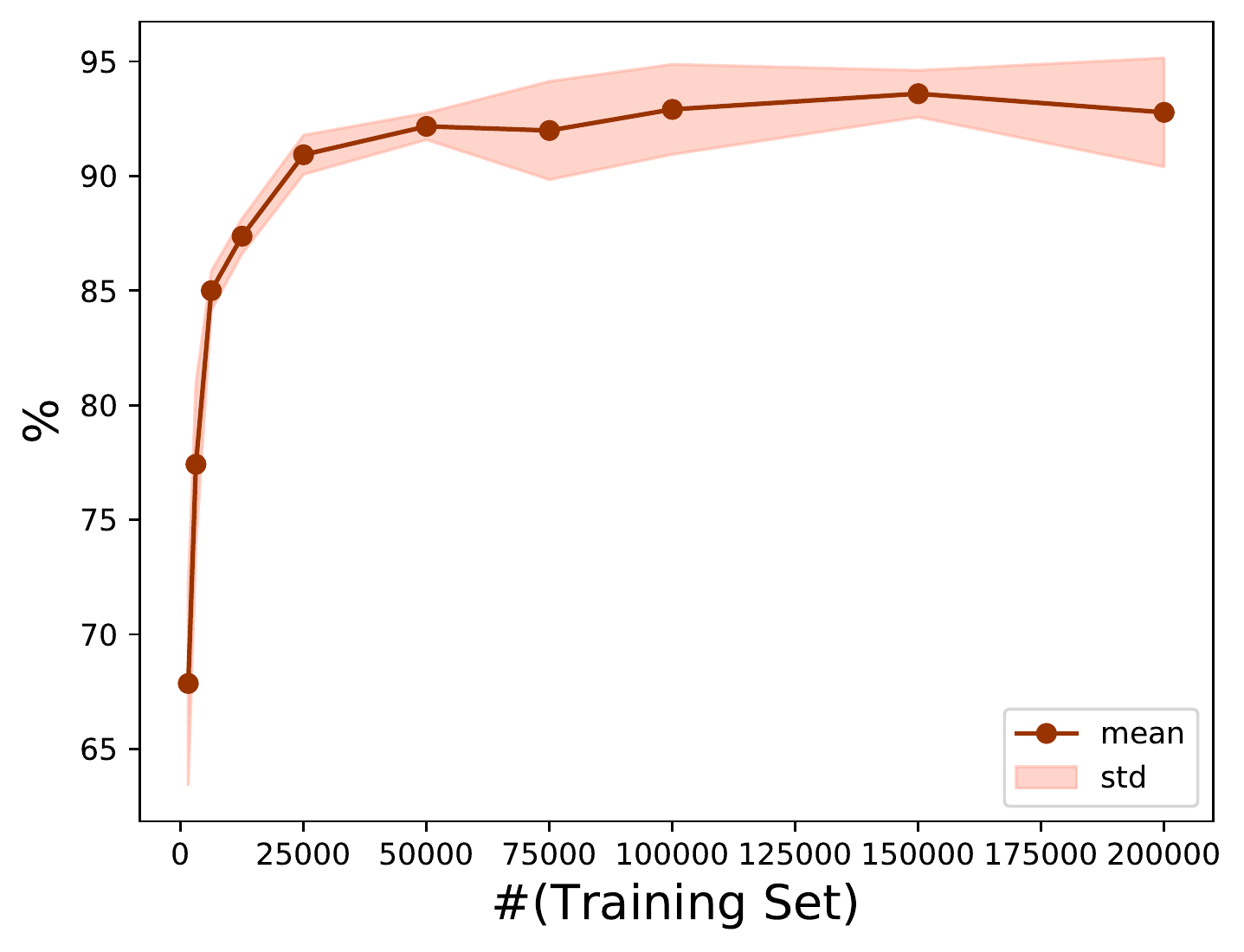}
\caption*{(b)}
\end{subfigure}
\caption{(a) MSE and the standard deviation for polygonal cylinders with $5$ random points. (b) Percentage of the unseen data and the standard deviation with absolute error less than $0.01$ as a function of the training set size.}
\label{fig:error-2}
\end{figure}

\begin{figure}[htp]
\centering
\includegraphics[width=.6\linewidth]{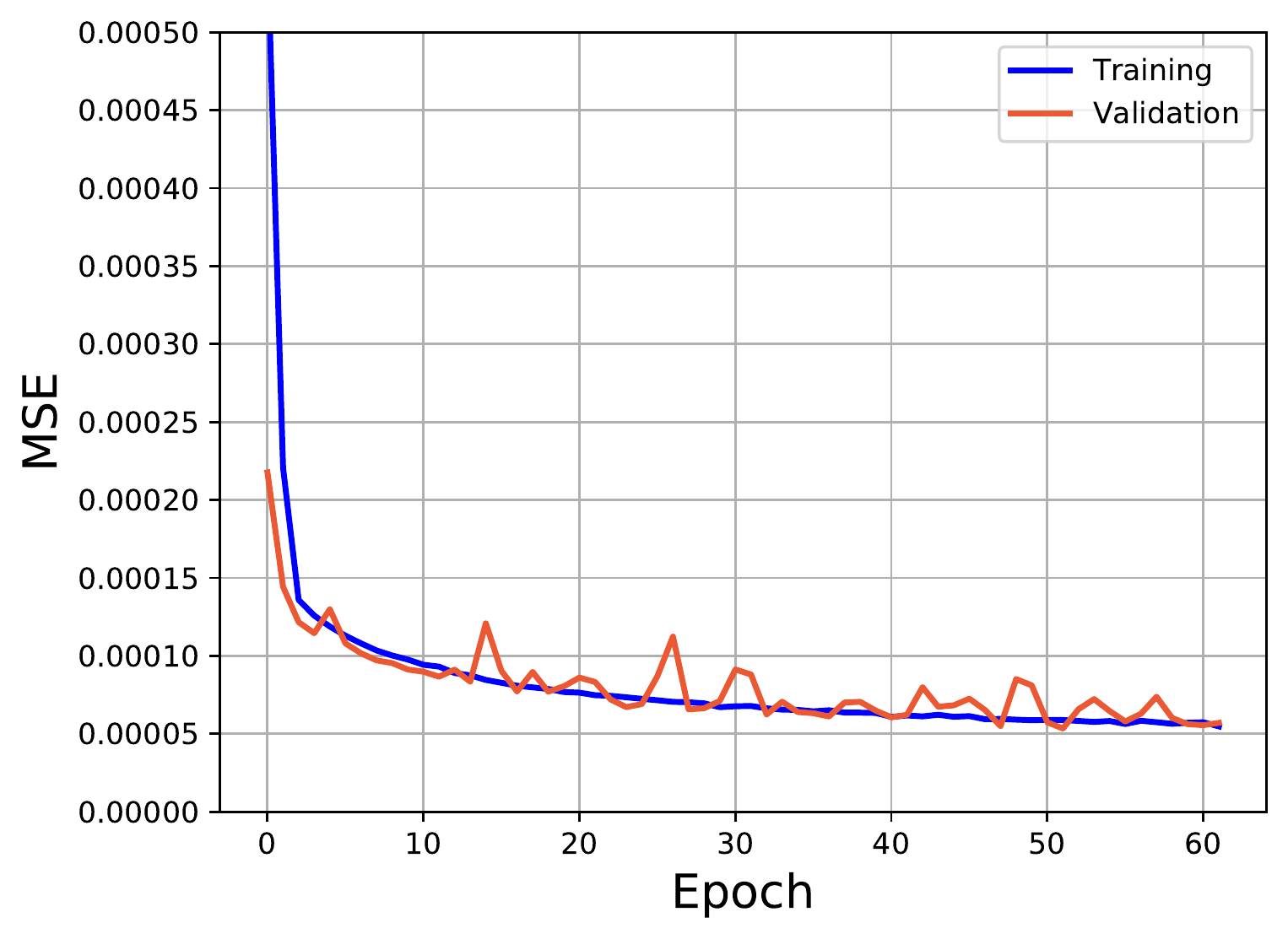}
\caption{The training history for $\Psi_{\mathrm{ml}}$ with the mean squared error for the total training set split into training and validation parts.
The history shown is for one trained model using a training set with 200\,000 points from the set Random 5.}
\label{fig:history}
\end{figure}

Since we present the mean-value of the MSE, we need to analyze the standard deviation.
In Figure~\ref{fig:error-2}, we present the MSE error and the percentage of unseen data with absolute error below threshold (red curve) together with the standard deviation for polygonal cylinders with five random points (shadowed region).  
Figure~\ref{fig:history} illustrates the MSE for different epoch numbers for training and validation sets.
We observe that the error on the training set, on the validation set, and on the test set are close.

When it comes to the choice of hyperparameters, the numerical experiments have shown that
reducing number of layers to two shows poor result, while increasing the number of layers and number of nodes in each layer does not improve much the result.
We have also tested YOGI~\cite{yogi}, but we did not manage to tune it to perform any better than ADAM.
The decaying learning rate gives better accuracy than a constant one.

\subsection{Performance on an out of sample set}
\label{sec:outofsample}

\begin{figure}[htp]
\centering
\includegraphics[width=.4\linewidth]{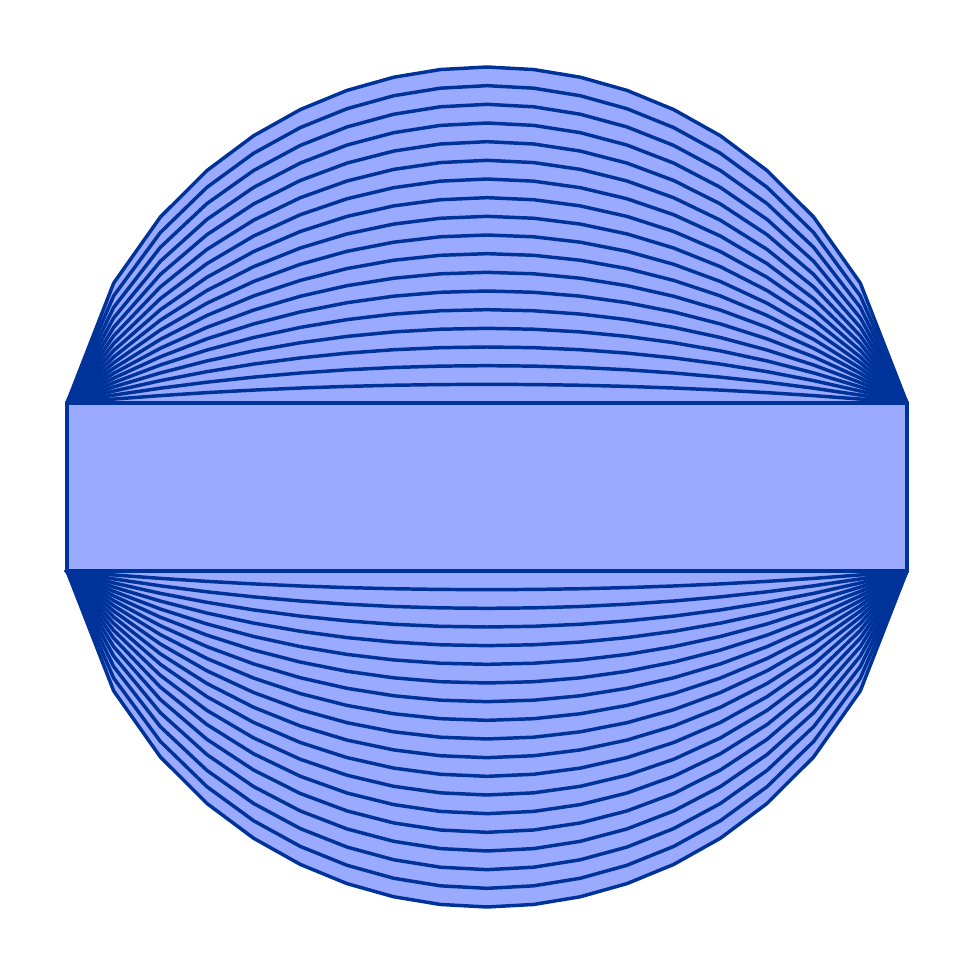}
\caption{The one-parameter out of sample family of convex cylinders.}
\label{fig:outofsamplepoly}
\end{figure}

To complement the evaluation of the trained models
on the unseen data, we here include an out of sample set.
Specifically, we choose a one-parameter family of 
convex symmetric cylinders defined as follows, and illustrated in Figure~\ref{fig:outofsamplepoly}.
For the model parameters of minimal and maximal radii 0.1 and 0.5, respectively, and center axis between 0 and 1, we let the midpoint radius $r(1/2)$ be a parameter varying between $0.1$ and $0.5$.
For each midpoint radius $r(1/2)$, we let $r(0) = r(1) = 0.1$ and construct a circular arc connecting the points $(0,0.1), (1/2,r(1/2)), (1,0.1)$.
In this way, by mirroring the arc, a symmetric convex cylinder is constructed.

\begin{figure}[htp]
\centering
\includegraphics[width=.6\linewidth]{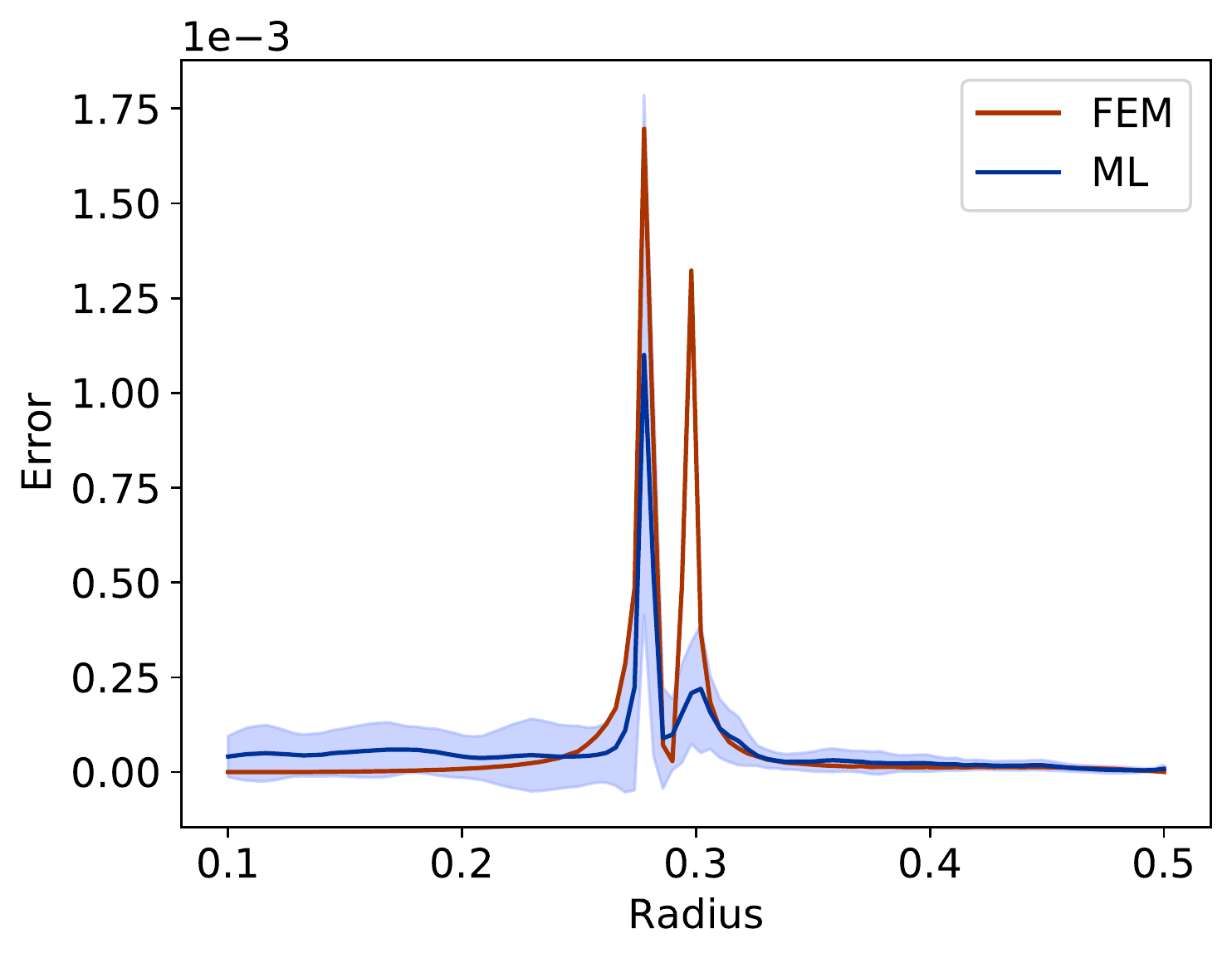}
\caption{The squared error between the 19 and 5 points FEM for radius $r(1/2)$,
and the mean squared error (shadowed area is standard deviation) in the predictions of the trained models ML. 
The models used here are the 10 obtained by training on sets with 200\,000 points from the set Random 5.
The two peaks come from that the 19 points cylinder and the 5 point cylinder, both hit the spectrum for exactly one value of $r(1/2)$ in the interval $[0.1,0.5]$, nonorthogonal to the data in $L^2$.}
\label{fig:outofsample}
\end{figure}

As justified by Lemma~\ref{lm:shape},
we may compute approximations of the averaged response $\Psi$,
using piecewise linear interpolations of the cylinder boundary in local charts.
We do this with $19$ uniform grid points on each arc,
as well as the down sampled $5$ point uniform grid arcs.
The squared error between 19 and 5 points numerically computed values of $\Psi$,
and the variation of the error over the parameter interval $[0.1,0.5]$ is shown in Figure~\ref{fig:outofsample} with label FEM.
The frequency range is again the interval $(\lambda_{\min}, \lambda_{\max}) = (0, 60)$.

We compute the predictions in $\Psi$ of the models trained with 200\,000 points from the set Random 5,
as evaluated in Figure~\ref{fig:error-1}.
In order to do so, we let the radius parameter $r(1/2)$ vary on a uniform grid of 100 points,
and down sample the cylinder radius to 5 points on each arc.
The mean squared error computed against the 19 points arc sets is shown in Figure~\ref{fig:outofsample} with label ML.

Numerically, 
the mean squared error in $\Psi$ between 19 points and the down sampled 5 points cylinder 
is truncated to $7.55 \cdot 10^{-5}$,
while the mean error of the predictions is truncated to $6.10 \cdot 10^{-5}$.
For comparison, we recall that the best DNN model evaluated to a mean squared error on the set Random 5 truncated to $5.50 \cdot 10^{-5}$, according to Table~\ref{tab:evamse}.
This verifies that indeed the performance of the trained models of the objective function $\Psi$
on sets of convex cylinders is indicated by the performance on our test sets when the domains are close, as described in Section~\ref{sec:shape}.

\section{Comparison with a linear model}
\label{sec:lin-vs-nonlin}

\begin{figure}[hbp!]
    \centering
    \begin{minipage}{.49\textwidth}
    \centering
    \includegraphics[width=\linewidth]{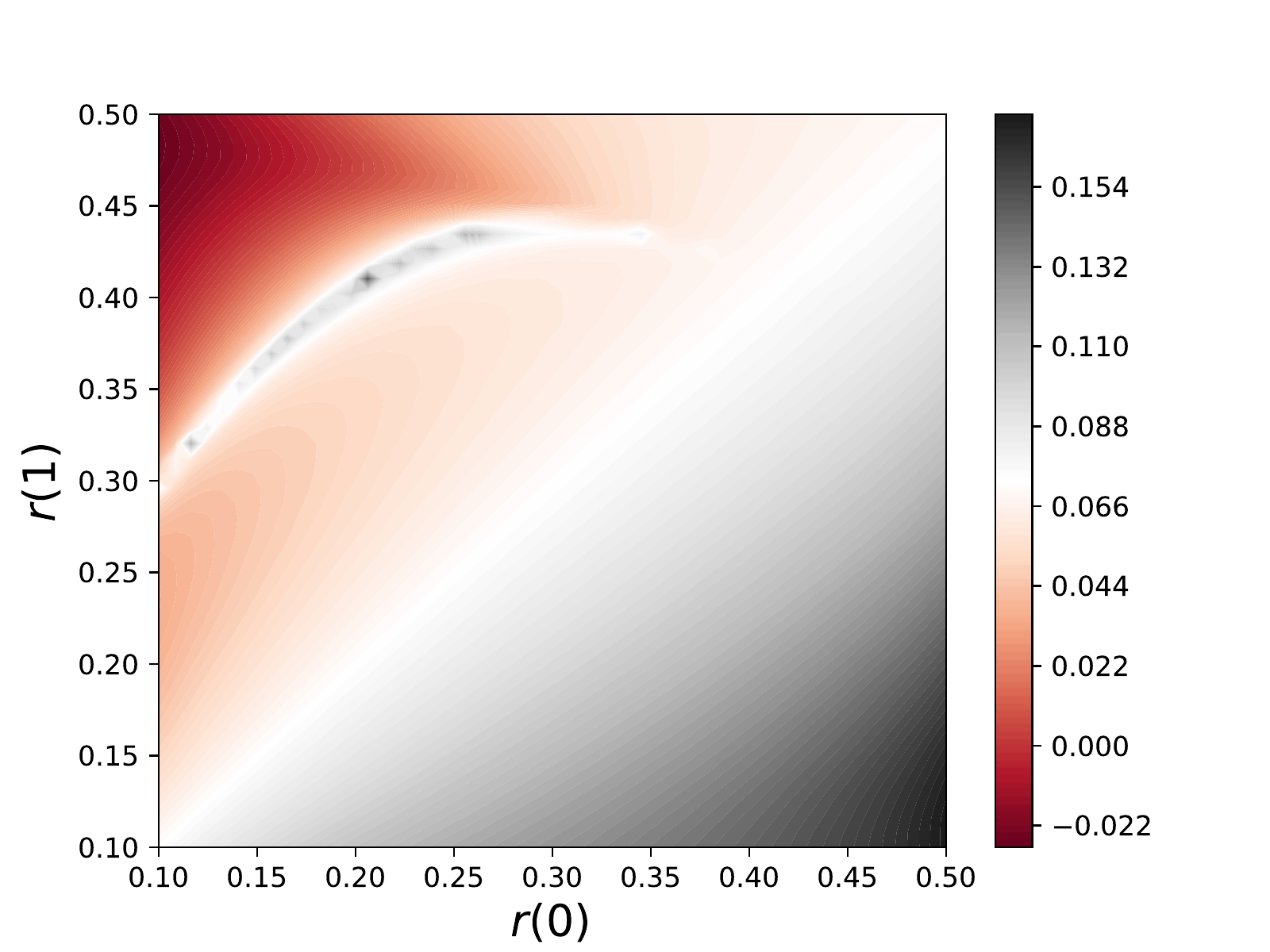}\\
    (a)
    \end{minipage}
    \begin{minipage}{.49\textwidth}
    \centering
    \includegraphics[width=\linewidth]{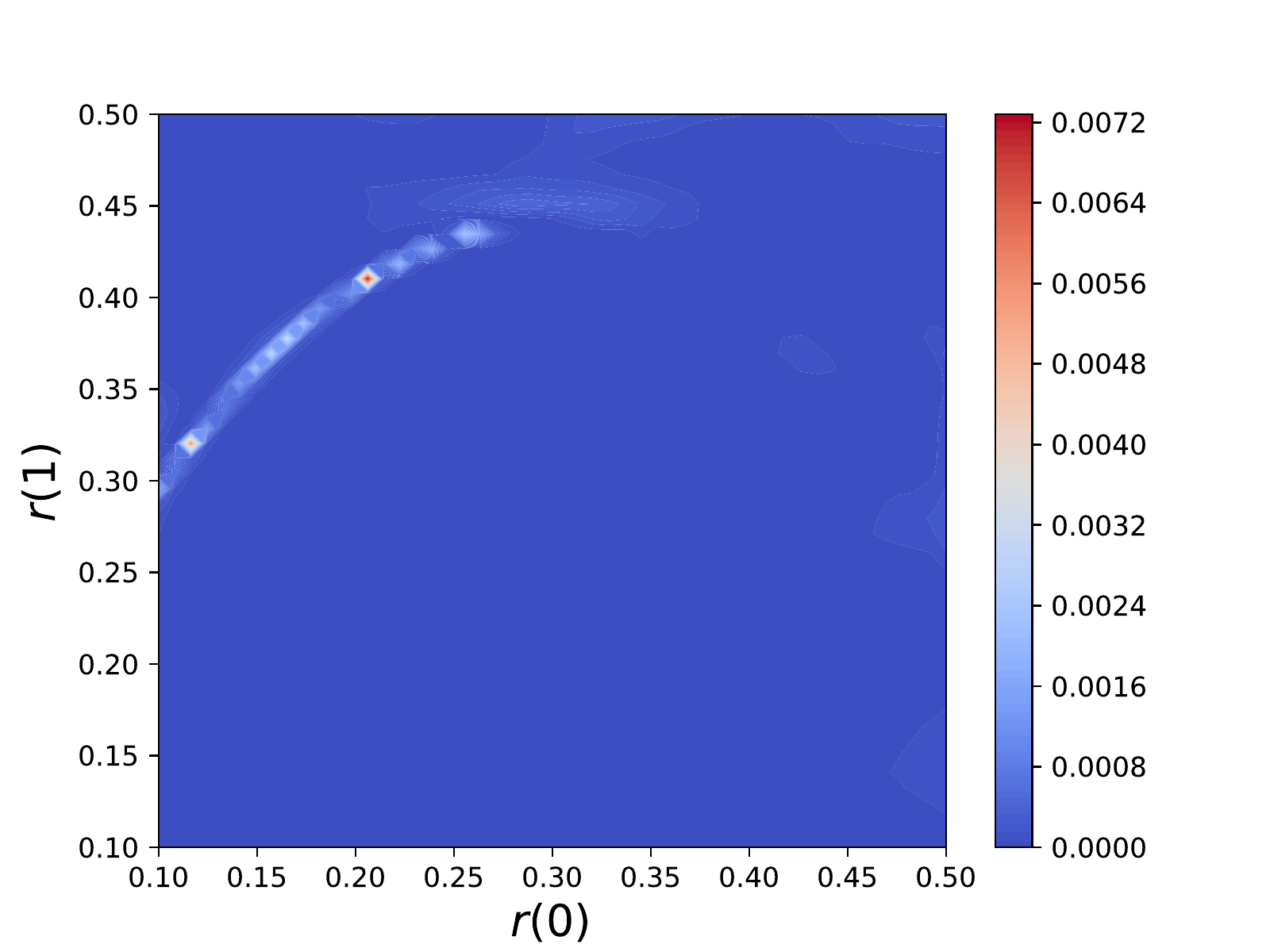}\\
    (b)
    \end{minipage}
    \caption{The value of $\Psi_h$ for cylinders with affine radii and $\lambda_{\max} = 60$ (a). The squared error between $\Psi_{\mathrm{ml}}$ and $\Psi_h$ (b).}
    \label{fig:unigrid2}
\end{figure}

One can ask why the linear regression would not be perform well in this case. 
To understand the nature of the nonlinearity in our problem,
we look at cylinders with radius $r(x_1)$ affine in $x_1$.
For radius $0.1 \le r \le 0.5$, and $x_1 \in (0,1)$, the set of cylinders 
may be parametrized by $r(0)$ and $r(1)$.
For the interval $(\lambda_{\min}, \lambda_{\max}) = (0, 60)$, 
we compute a numerical approximation $\Psi_h$ of the objective 
function $\Psi$.
In Figure~\ref{fig:unigrid2}(a), $\Psi_h$ is shown.
Of course $\Psi$ is linear for uniform cylinders, as it is constant.
On the diagonal $r(0) = r(1)$ in the Figure~\ref{fig:unigrid2}(a), we see the value of this constant.
Off the diagonal, we see that $\Psi_h$ is clearly not the graph of a linear
function.
A more careful inspection shows that $\Psi_h$ is 
smooth and seems to be linear everywhere except in the upper left corner of the
figure, where it shows rapid growth in a narrow region.
We know that $\Psi$ and $\Psi_h$ are defined for intervals $(0,\lambda_{\max})$ for almost every
$\lambda_{\max} > 0$, but not for all. Namely,
$\Psi_h$ might show singular behavior in the
vicinity of a set of positive one-dimensional measure, where $p_\lambda$ is singular. The approximation $\Psi_{\mathrm{ml}}$ we get with the ML algorithm gives largest error exactly in this singularity region, as seen in Figure~\ref{fig:unigrid2}(b). 
This verifies the need for a nonlinear activation function in our problem
even if restricted to the cylinders with affine radii as functions of $x_1$.

In Tables~\ref{tab:evamse} and \ref{tab:evaaethr}, the errors of the approximations 
on the Uniform and Random 5 test sets of unseen data are provided.
The floating point numbers have been truncated.
For comparison, we include both a linear model and the proposed nonlinear model DNN.
We train the models for polygonal cylinders with five random points. One can see that the proposed DNN model performs much better than the linear one, both on uniform and non-uniform cylinders.

\begin{table}[htp!]
\centering
\begin{tabular}{llrrr}
 & \multicolumn{2}{c}{\textbf{TS = Training Set}}  & \multicolumn{2}{c}{\textbf{Mean Squared Error}} \\[2mm]
\textbf{Model} & \textbf{TS}         & \textbf{\#(TS)}        & \textbf{Uniform} & \textbf{Random 5}  \\
Linear & Random 5 & 200\,000   &  2.56e-3  &  1.91e-3  \\
DNN    & Random 5 & 200\,000  &  2.31e-5  &  5.50e-5\\[2mm]
\end{tabular}
\caption{The performance on unseen data of the converged linear and nonlinear models.
The mean squared error on unseen data.
For the dense neural network model (DNN), the performance is for the best of the sampled models in the sense of minimum mean squared error.
The floating point numbers are truncated.}
\label{tab:evamse}
\end{table}

\begin{table}[htp!]
\centering
\begin{tabular}{llrrr}
 & \multicolumn{2}{c}{\textbf{TS = Training Set}}  & \multicolumn{2}{c}{\textbf{\% Abs Err $\mathbf{< 0.01}$}} \\[2mm]
\textbf{Model} & \textbf{TS}         & \textbf{\#(TS)}        & \textbf{Uniform} & \textbf{Random 5}  \\
Linear & Random 5 & 200\,000  & 46.9 & 22.1 \\
DNN    & Random 5 & 200\,000 & 98.3 & 95.7 \\[2mm]
\end{tabular}
\caption{The performance on unseen data of the converged linear and nonlinear models.
The percentage of unseen data with an absolute error $|\Psi_\mathrm{ml} - \Psi_h|$ less than $0.01$).
For the dense neural network model (DNN), the performance is for the best of the sampled models in the sense of minimum mean squared error.
The floating point numbers are truncated.}
\label{tab:evaaethr}
\end{table}

\section{Conclusions}
\label{sec:conclusions}
We have proposed a feedforward dense neural network for predicting the average sound pressure response over a frequency range.
We have shown for polygonal cylinders that the obtained results are sufficiently accurate in that they reach the estimated accuracy of the numerical data.
Although the amount of data needed in order to reach the desired accuracy could be considered as big, it is expected that the results would serve as a point of reference for more advanced machine learning models.
The performance of the feedforward dense neural network has been evaluated.
The dependence of the percentage accurately predicted samples and the mean squared error on the training set size is presented.

\subsection*{Acknowledgments.}
The computations were partially performed on resources at Chalmers Centre for Computational Science and Engineering (C3SE) provided by the Swedish National Infrastructure for Computing (SNIC).
We thank the referees for careful reading of the paper.

\clearpage

\appendix

\section{A Hyperparameter grid}
\label{app:hyper}

To supplement Section~\ref{sec:hyper} we here provide a 
grid around the model and training parameter values specified.
In particular, we choose the parameters (i) number of hidden layers, (ii) number of nodes in each layer, and (iii) the fraction of the training data used for the validation split.
For each 3-tuple of parameters, we present in Table~\ref{tab:hyperparameter} the mean squared error on unseen data from the data sets Uniform and Random 5.
The data sets used for training was 200\,000 points from the Random 5 training set.

\begin{table}[htp!]
\centering
\begin{tabular}{ccccc}
\multicolumn{3}{c}{\textbf{Hyperparameter}}
& \multicolumn{2}{c}{\textbf{Mean Squared Error}} \\[2mm]
\textbf{Hidden Layers} & \textbf{Layer Size} & \textbf{Validation Split \%}        & \textbf{Uniform} & \textbf{Random 5}  \\
2 & 64 & 10  & 3.98e-5 & 1.17e-4 \\
2 & 64 & 20  & 4.64e-5 & 1.53e-4 \\
2 & 64 & 30  & 1.37e-4 & 1.44e-4 \\
2 & 128 & 10 & 4.80e-5 & 8.45e-5 \\
2 & 128 & 20 & 5.00e-5 & 8.74e-5 \\
2 & 128 & 30 & 4.65e-5 & 9.61e-5 \\
2 & 192 & 10 & 1.21e-5 & 7.38e-5 \\
2 & 192 & 20 & 3.15e-5 & 7.98e-5 \\
2 & 192 & 30 & 9.42e-6 & 8.47e-5 \\
3 & 64 & 10  & 5.21e-5 & 8.25e-5 \\
3 & 64 & 20  & 1.79e-5 & 7.91e-5 \\
3 & 64 & 30  & 4.44e-5 & 9.35e-5 \\
3 & 128 & 10 & 1.78e-5 & 6.01e-5 \\
3 & 128 & 20 & 1.29e-5 & 5.75e-5 \\
3 & 128 & 30 & 9.45e-5 & 5.75e-5 \\
3 & 192 & 10 & 1.75e-5 & 5.09e-5 \\
3 & 192 & 20 & 1.07e-5 & 5.20e-5 \\
3 & 192 & 30 & 7.97e-6 & 5.55e-5 \\
4 & 64 & 10  & 2.02e-5 & 7.40e-5 \\
4 & 64 & 20  & 1.86e-5 & 6.26e-5 \\
4 & 64 & 30  & 2.47e-5 & 8.25e-5 \\
4 & 128 & 10 & 1.68e-5 & 5.34e-5 \\
4 & 128 & 20 & 3.00e-5 & 5.60e-5 \\
4 & 128 & 30 & 3.31e-5 & 5.46e-5 \\
4 & 192 & 10 & 1.04e-5 & 4.91e-5 \\
4 & 192 & 20 & 1.40e-5 & 5.39e-5 \\
4 & 192 & 30 & 1.80e-5 & 5.27e-5 \\[2mm]
\end{tabular}
\caption{
The mean squared errors on the unseen data Uniform and Random 5
for various values of the parameters.
The numerical values of the mean squared errors are truncated.}
\label{tab:hyperparameter}
\end{table}

\clearpage

\bibliographystyle{plain}
\bibliography{refs}

\end{document}